\documentclass{article}

\usepackage{arxiv}

\usepackage[utf8]{inputenc} 
\usepackage[T1]{fontenc}    
\usepackage[hidelinks]{hyperref}
\usepackage{url}            
\usepackage{booktabs}       
\usepackage{amsfonts}       
\usepackage{amsmath}
\usepackage{amssymb}
\usepackage{nicefrac}       
\usepackage{microtype}      
\usepackage{cleveref}       
\usepackage{lipsum}         
\usepackage{graphicx}
\usepackage{natbib}
\usepackage{doi}
\usepackage{lineno}         
\usepackage{fontawesome5}

\title{Towards Generalizable Mapping of Hedges and Linear Woody Features from Earth Observation Data: a national Product for Germany}

\newif\ifuniqueAffiliation

\ifuniqueAffiliation 
\else
\usepackage{authblk}

\newbox{\orcid}\sbox{\orcid}{\includegraphics[scale=0.06]{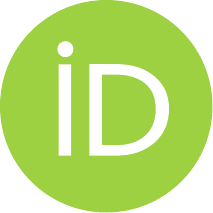}} 
\author[1]{%
	\href{https://orcid.org/0000-0002-7179-3664}{\usebox{\orcid}\hspace{1mm}Thorsten Hoeser}%
	\thanks{Corresponding author \texttt{thorsten.hoeser@dlr.de}}%
}

\author[1]{%
	\href{https://orcid.org/0009-0007-0097-2714}{\usebox{\orcid}\hspace{1mm}Verena Huber-García}%
}

\author[1]{%
	\href{https://orcid.org/0000-0002-7302-6813}{\usebox{\orcid}\hspace{1mm}Sarah Asam}%
}

\author[1]{%
	\href{https://orcid.org/0000-0002-8221-2554}{\usebox{\orcid}\hspace{1mm}Ursula Gessner}%
}

\author[1,2]{%
	\href{https://orcid.org/0009-0007-4933-5898}{\usebox{\orcid}\hspace{1mm}Claudia Kuenzer}%
}

\affil[1]{Earth Observation Center (EOC), German Aerospace Center (DLR), Oberpfaffenhofen, 82234 Wessling, Germany}

\affil[2]{Institute for Geography and Geology, University of Wuerzburg, 97074 Wuerzburg, Germany}
\fi

\renewcommand{\headeright}{ }
\renewcommand{\undertitle}{Preprint}

\hypersetup{
pdftitle={Towards Generalizable Mapping of Linear Woody Features from Heterogeneous Earth Observation Data},
pdfauthor={Thorsten~Hoeser, Verena~Huber-García, Sarah~Asam, Ursula~Gessner, Claudia~Kuenzer},
}

\begin{document}
\maketitle
\begin{abstract}
Hedges and other linear woody features provide valuable ecosystem services, particularly within intensively managed agricultural landscapes. They are key elements for climate adaptation and biodiversity amongst others not only due to a largely varying flora, but also as a feeding-, resting-, and nesting place for many animals and insects including valuable pollinators. Therefore, they require dedicated management, preservation, and attention. Thus, systematic and large-scale mapping of these features from Earth observation data is of high importance. However, transferable and reusable workflows for linear woody feature mapping remain a key methodological challenge, given the diversity of sensor types, spatial resolutions, data acquisition conditions, and complex landscape variability encountered across study areas. We introduce a modular workflow built around two independently optimizable components. Firstly, a flexible input data interface that consolidates heterogeneous Earth observation data into a binary woody vegetation mask, and secondly, a deep neural network trained to separate linear from non-linear shapes within these masks. We demonstrate the workflow by deriving three national-scale linear woody feature maps for all of Germany from three input sources with 0.73~m, 1~m and 3~m spatial resolution, respectively, by using a single trained model without retraining. Evaluation against refined reference data from four federal state biotope mapping campaigns and comparison with two existing linear woody feature maps demonstrate that the workflow produces competitive results across all evaluation sites on a national level. The modular design and its demonstrated applicability at national scale provide a foundation for scalable and generalizable linear woody feature mapping beyond Germany.
\end{abstract}



\section{Introduction}
Woody features are essential components of open landscapes as they provide a wide range of ecosystem services. They are important habitats for a large variety of floral and faunal species, they reduce soil erosion by water and deflation by wind, as well as evapotranspiration, and increase water retention \citep{XIAO2024111438, HOLDEN20191, GHAFARIAN2024109949, KRATSCHMER202428, HOLLAND2000115, Brandle2004}. Further, they store significant amounts of above and belowground carbon \citep{ Marcantonio2024, DREXLER2024116878, Drexler2021} and are, in several regions, of high value for recreation and cultural landscape identity \citep{BUREL1995327, su17219865}. The conservation, regrowth, sustainable management and thus also the monitoring of such woody landscape features are receiving increasing attention from diverse stakeholder groups, including ecologists and agricultural practitioners, and are promoted in several legislations such as the EU Nature Restoration Regulation \citep{EU2024NatureRestoration}, the EU Common Agricultural Policy \citep{EuropeanCommissionCAP}, and the EU Biodiversity Strategy \citep{doi/10.2779/677548}. Among the woody features in open landscapes, linear woody features, such as hedgerows, tree lines, and riparian vegetation, play an important role. Compared to more compact woody features, their elongated shape makes them particularly effective at reducing runoff and hindering deflation of soil, while also enhancing their habitat value by providing ecological corridors and extensive edge habitats \citep{Montgomery2020hedgerows, GARRATT2017363, PELLETIERGUITTIER2020107079}. Ecotones exhibit higher biodiversity and ecological complexity than adjacent habitats alone, serving as refuges and corridors for species that depend on environmental conditions of both ecosystems, a phenomenon well-documented in studies on landscape ecology \citep{KRATSCHMER202428, klimm2025}. Further, they have traditionally been an integral part of cultural landscapes, as they naturally grow along field boundaries, streets, railroads, waterways, and natural water bodies.

Earth observation (EO) has become a key tool to derive large-scale inventories of woody cover, enabling the systematic mapping of its spatial distribution and temporal dynamics. The separation of woody landscape elements however is less evolved, with the Small Woody Features (SWF) layer published by the European Union’s Copernicus Land Monitoring Service (CLMS) having been the only routinely produced EO data set in this domain \citep{Faucqueur2019swf}. The availability of high-resolution EO data, combined with automated analysis methods, now enables efficient large-area assessments. This is particularly relevant because manual digitization of woody features is labour-intensive and becomes infeasible when addressing large spatial extents or when frequent updates are required.

Mapping linear woody features using EO data has therefore been the focus of numerous studies across different spatial scales, sensor types, and application contexts. In addition to older studies like the ones from \citet{5256249, 6723533, rs6053752, TANSEY2009145, 5651636, 6947596, OCONNELL2015165}, several recent studies on the detection of linear woody features exist: \citet{HUBERGARCIA2025101451} mapped hedgerows across Bavaria, Germany, from orthophotos with a 20~cm spatial resolution using the DeepLabV3 \citep{chen2018encoderdecoder}, a Convolutional Neural Network architecture. A preliminary study of the Bavarian-wide product was carried out by \citet{10.1117/1.JRS.15.018501} who detected hedgerows for one study site in Bavaria from IKONOS imagery at 1~m spatial resolution testing several neural networks. \citet{MURO2025114870} made use of a U-Net architecture for the segmentation of hedgerows from PlanetScope data with 3~m spatial resolution acquired between April and October 2022 at the national scale of Germany.

There are other regional to national studies across Europe such as \citet{Wolstenholme2025automated} who identified hedgerows and hedgerow gaps from 25~cm resolution aerial imagery using the U-Net architecture for a region in the UK, while \citet{BROUGHTON2025126705} modelled linear woody features for entire UK using a rule-based approach from 1~m resolution LiDAR data. In another study by \citet{conserva2025mapping} covering the UK, the authors mapped hedges, woodlands as well as stone walls using a deep learning model on 25~cm resolution orthophotos. \citet{rs17091506} focused on the semi-automatic extraction of hedgerows by applying Object-Based Image Analysis (OBIA) to PlanetScope (3~m resolution) and Sentinel-2 (10~m resolution) in Northern Italy. \citet{GRONDARD2025113853} monitored woody landscape features for parts of the Netherlands following a rule-based approach and making use of 50~cm resolution LiDAR data. Finally, in the study by \citet{rs15112766} a U-Net-based segmentation model was used for detecting woody vegetation landscape features including hedges, tree lines, and riparian vegetation from aerial photography for a study region in Slovenia.

Besides, there are several studies focusing on trees outside forests (ToF) which include also linear woody features to a substantial extent. \citet{Meneguzzo2013} compared two approaches, OBIA and a pixel-based classification approach, to map ToF for a small study area in the USA from 1-2~m resolution aerial imagery. OBIA was also used by \citet{pujar2014} to estimate ToF from a combination of panchromatic Cartosat~1 data (2.5~m resolution) and multispectral LISS data (5.8~m resolution) for a study area in India. \citet{MAACK2017118} applied a rule-based approach to derive ToF for several study areas in Germany from a LiDAR-derived normalized Digital Surface Model (nDSM) at a 0.5-1~m resolution. More recently, \citet{doi:10.1126/sciadv.adh4097} used a deep learning approach on PlanetScope data at 3~m resolution to map ToF over entire Europe, however not explicitly identifying hedgerows. \citet{lucas2025mapping} mapped and classified ToF into the classes forest, patch, linear and tree for several study areas across Germany using a deep learning approach on 0.2~m resolution aerial imagery.

There are further studies mapping single tree crowns excluding smaller bushes such as \citet{Brandt2020} and \citet{Tucker2023} who detected tree crowns across semi-arid sub-Saharan Africa north of the Equator. They relied on machine learning techniques and used multispectral data from several sensors QuickBird-2, GeoEye-1, WorldView-2 and WorldView-3 with a 0.5~m spatial resolution. Also \citet{Reiner2023} mapped the tree crown cover, both within and outside forests, for entire Africa using PlanetScope data and a U-Net architecture.

\begin{figure}[htbp]
\centering
\includegraphics[width=\textwidth]{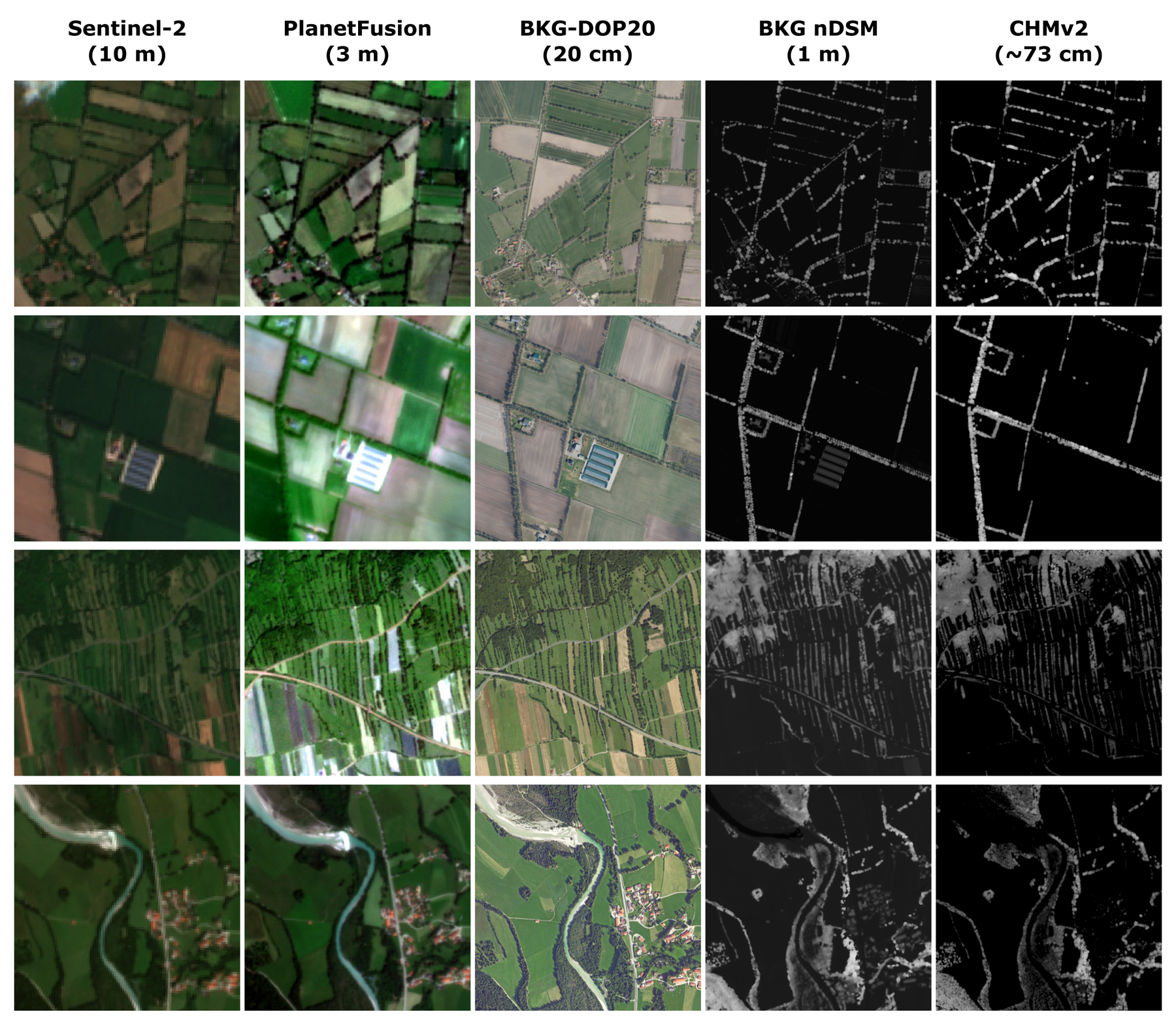}
\caption{Comparison of different sensor and elevation data products across different sites in Germany, ranging from north (top) to south (bottom), showing different configurations in linear woody feature morphologies and their representation at different spatial resolution. DOP20: Geobasisdaten: ©~GeoBasis-DE~/~BKG (2023)}
\label{fig:diff_eo_data}
\end{figure}

However, these efforts often differ substantially in their spatial resolution and extent, landscape type and definitions of what constitutes a linear woody feature. As a result, many existing approaches are highly specialized, relying on narrowly optimized models and workflows tailored to particular combinations of input data configuration like a specific sensor, certain acquisition times, and confined study areas, see Figure~\ref{fig:diff_eo_data} for example differences. While such approaches can achieve high performance \citep{HUBERGARCIA2025101451, lucas2025mapping}, transferring them to different regions, data sources, or conditions of acquisition often leads to significant performance degradation or may require extensive retraining and workflow adjustments.

In this study, we introduce an approach designed to address these limitations by aiming for sensor-independent and data-flexible extraction of linear woody features. The proposed workflow is capable of incorporating heterogeneous types of EO data and EO-derived products as input sources and translating them to an intermediate binary woody vegetation mask before linear woody feature extraction. Furthermore, the approach emphasizes spatial transferability by employing a training strategy that does not rely directly on site-specific real-world labels but uses synthetically generated, and thus site-independent, training data. 

We present this approach for linear woody feature extraction, its application, and potential for methodological refinement. To demonstrate its flexible applicability, we generate three national-scale linear woody feature maps for all of Germany derived from three different data sources, each representing distinct EO sensor types and spatial resolutions. Our overarching objective is to move beyond narrowly defined data-site-model-application configurations and instead establish a consolidated yet flexible and thus transferable workflow for linear woody feature mapping in general. At the core of the developed workflow is a deep learning-based segmentation model trained to disentangle non-linear or patchy woody vegetation from linear woody vegetation, combined with an input interface that allows heterogeneous data sources, including EO imagery, EO-derived products, and GIS data sets, to be integrated individually or jointly. We further perform a comparative evaluation for the three data sets and two existing data sets from Germany, and discuss advantages and disadvantages of the proposed workflow, as well as options for future research.

The main contributions of this study are as follows:
\begin{itemize}
\item A conceptual workflow for linear woody feature separation, enabling heterogeneous input data sets to be integrated into a consolidated processing pipeline.
\item A training methodology for a spatially independent linear woody feature separator based on automated, synthetic training data generation.
\item Three national-scale linear woody feature maps for Germany, produced at spatial resolutions ranging from 0.73 to 3~meters.
\item An evaluation based on independent reference data sets and comparison
 of the derived linear woody feature maps against two existing linear woody feature maps in Germany.
\end{itemize}

\section{Data and Materials}
\subsection{Input data for linear woody feature mapping}\label{sec:input_data}

\begin{figure}[htbp]
\centering
\includegraphics[width=\textwidth]{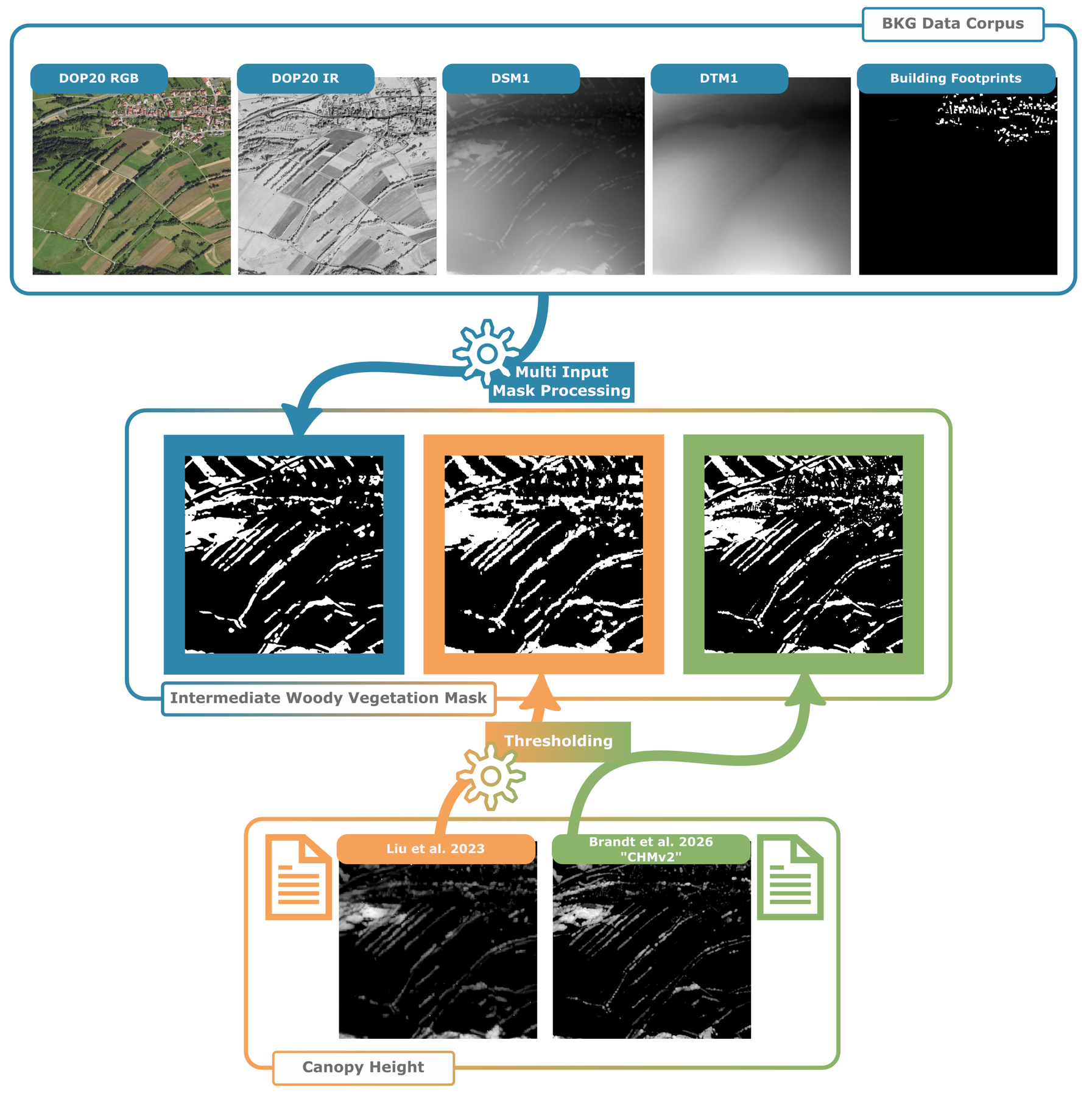}
\caption{Overview of the three different input sources separated into a set of heterogeneous data layers provided by the BKG (top), and canopy height map input data (bottom row) \citep{doi:10.1126/sciadv.adh4097, brandt2026chmv2}. All input data sets are used to derive a consolidated intermediate data product, a binary woody vegetation mask, one for each input data set (middle). DOP20, DTM1, DSM1: Geobasisdaten: ©~GeoBasis-DE~/~BKG (2023)}
\label{fig:data}
\end{figure}

Data used in this study to generate binary woody feature maps and subsequently separate linear woody features, originate from three different sources, see Figure \ref{fig:data}. The first source is a heterogeneous data corpus provided by the Federal Agency for Cartography and Geodesy of Germany (Bundesamt für Kartographie und Geodäsie, BKG), which includes digital orthophotos with a spatial resolution of 20~cm (DOP20) and four spectral bands (RGB and NIR) \citep{BKG2026DOP20}. In addition, the BKG provides a digital surface model (DSM1) \citep{BKG2026DOM1} and a digital terrain model (DTM1) \citep{BKG2026DGM1, AdV2025PQS_DGM}, each with a spatial resolution of 1~m. BKG further supplies vector data representing building footprints across Germany for the year 2023. The national scale data sets of the BKG are typically provided in EPSG:25832 and the raster data is following a 1 by 1~km tile grid.

Germany is organized into 16 federal states. Due to this political structure, the nationwide DOP20, DSM1, and DTM1 mosaics used in this study are compiled by the BKG from data sets provided by these 16 federal states individually. While the DSM1 and DTM1 products are delivered with harmonized value ranges and include explicit efforts to ensure consistency across state boundaries \citep{BKG2026DOM1}, the DOP20 mosaic is designed to serve as a high-quality visualization layer. This intended purpose introduces several challenges when the data set is used for analytical applications, as the airborne data is often times acquired over different years, during different months, sometimes during leaf-off or leaf-on season, with different sensors.

\begin{figure}[htbp]
\centering
\includegraphics[scale=0.75]{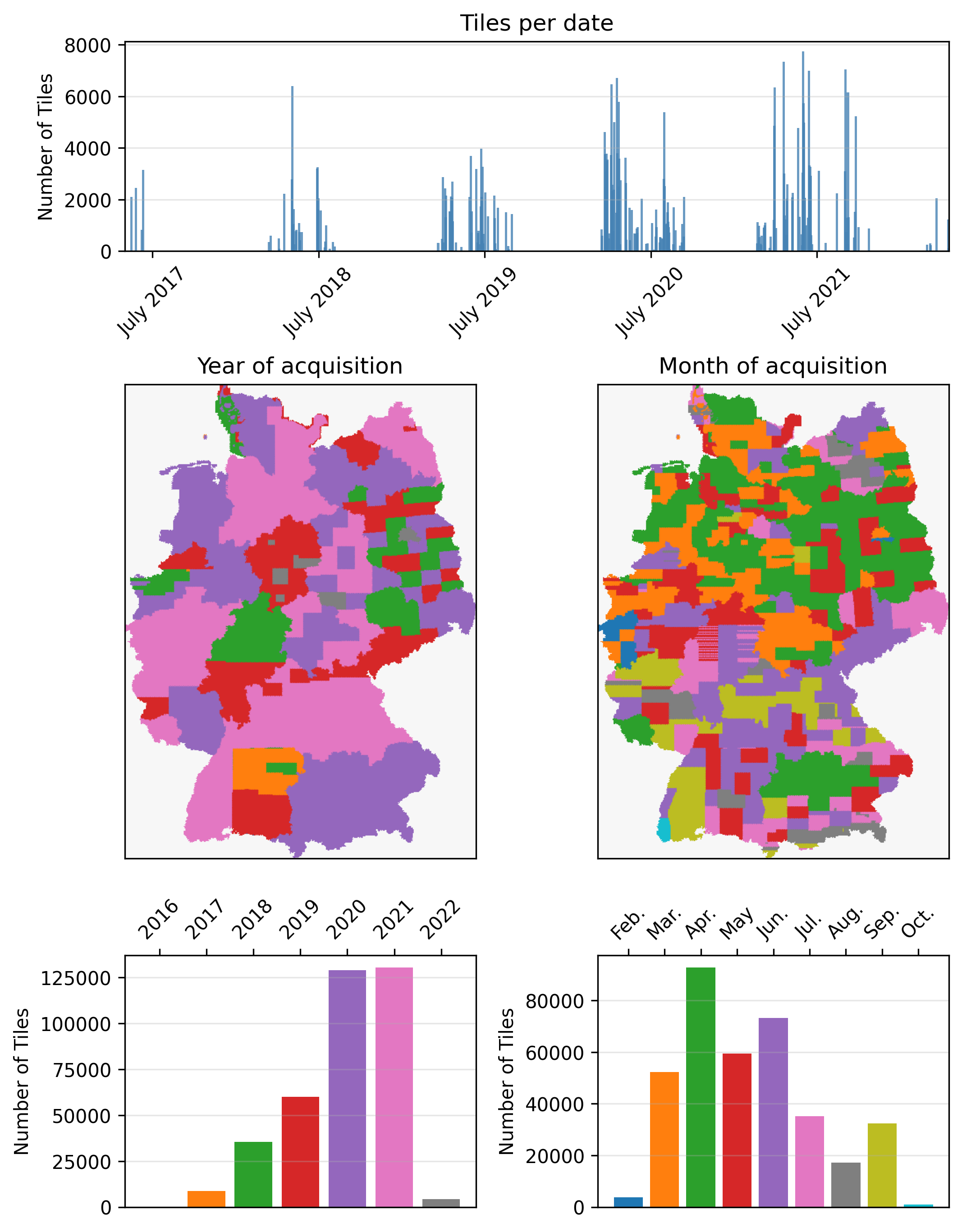}
\caption{Overview of the acquisition dates of the DOP20 BKG mosaic tiles and their spatial distribution across Germany.}
\label{fig:meta_I_bkg}
\end{figure}

\begin{figure}[htbp]
\centering
\includegraphics[scale=0.75]{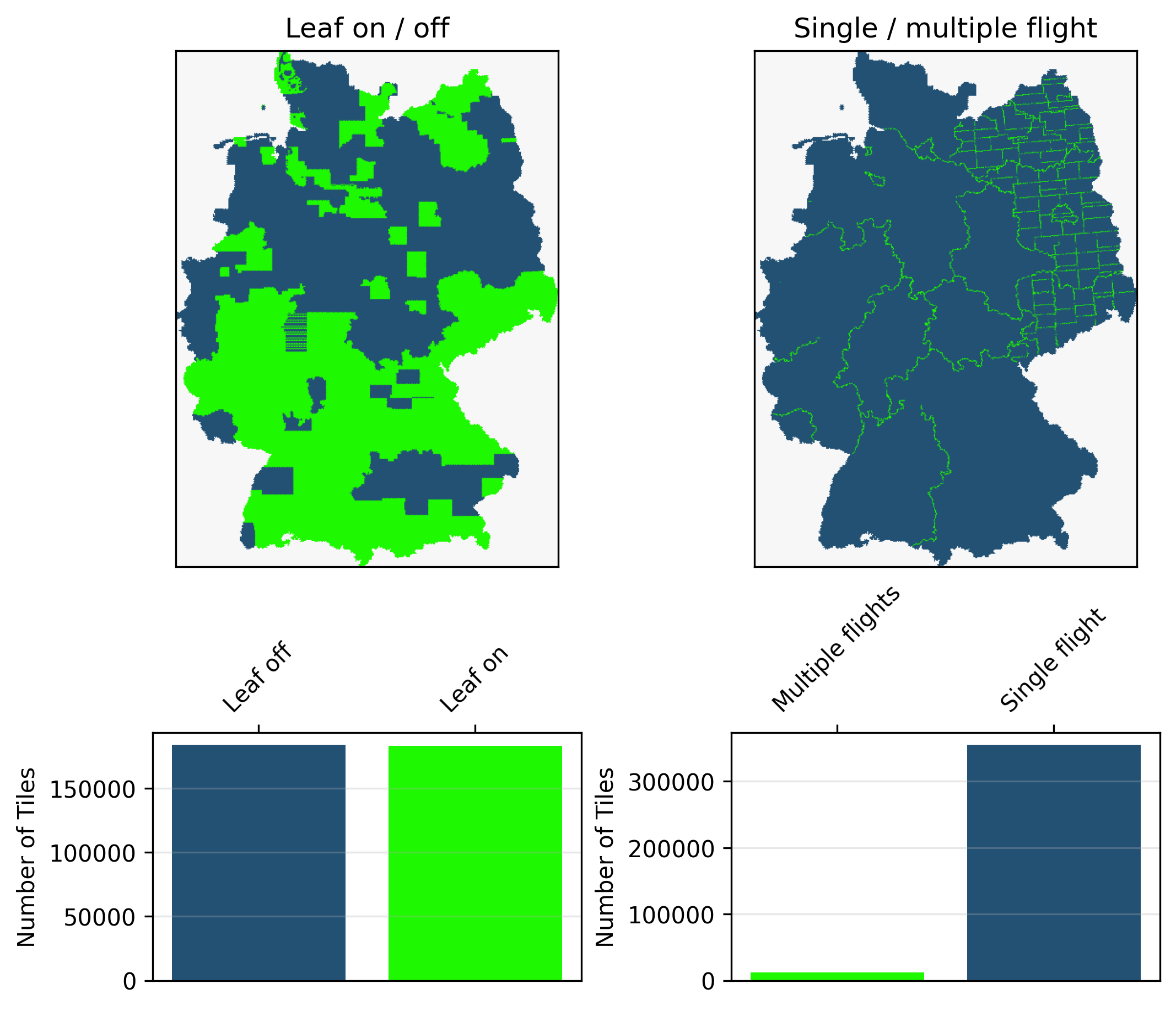}
\caption{Overview of the spatial distribution of tiles of the BKG DOP20 mosaic and their leaf on / off condition, where leaf on tiles acquired between April 15 and September 15, and tiles with data coming from a single or multiple flight campaigns.}
\label{fig:meta_II_bkg}
\end{figure}

Figure~\ref{fig:meta_I_bkg} illustrates the spatial and temporal distribution of acquisition dates contributing to the DOP20 mosaic. The revealed patterns align with state borders, particularly at an annual scale, reflecting typical update cycles across the federal states. Additionally, a pronounced north-south gradient emerges at the level of acquisition months. Northern regions tend to acquire imagery during leaf-off conditions early in the year, whereas southern regions more commonly capture data during leaf-on periods in mid-year, see also Figure~\ref{fig:meta_II_bkg}.

As the BKG distributes the DOP20 mosaic in tiles of 1,000 $\times$ 1,000~m, individual tiles may contain imagery derived from multiple flight campaigns. Because acquisition dates are provided at the tile level, it is not always possible to unambiguously assign specific pixels within such tiles to a precise acquisition date, see Figure \ref{fig:meta_II_bkg}.

Furthermore, in line with its purpose as a visualization layer, the digital numbers of the DOP20 mosaic are contrast-enhanced through scaling procedures, while those are not provided alongside the mosaic tiles. In addition, the tiles are provided as JPEG compressed 8-bit tif files with different factors for RGB and IR channels \citep{BKG2026DOP20}. Therefore, global or tile-based procedures to convert the digital numbers to surface reflectance values are not straightforward. The resulting properties have to be accounted for, especially for vegetation analysis, where factors such as varying acquisition phenology like leaf-off conditions and inconsistencies in radiometric scaling across the mosaic must be carefully addressed. Despite these limitations, the geometric accuracy and orthorectification quality of the DOP20 mosaic make it a valuable data source with fine grained textural features.

The two other data sources used in this study are canopy height maps derived from EO data. The data product presented by \citet{doi:10.1126/sciadv.adh4097} uses a deep learning approach trained on airborne LiDAR data to create a binary canopy / no-canopy mask, and a regression model to predict canopy height within this mask across Europe finally derived from PlanetScope imagery. The final canopy height map has a spatial resolution of 3~m, corresponding to the resolution of the underlying PlanetScope imagery acquired in 2019.

The second canopy height map used as input in this study is the Canopy Height Map v2 (CHMv2), introduced by \citet{brandt2026chmv2}, which is the successor of the canopy height map by \citet{TOLAN2024113888}. CHMv2 is generated using a deep learning model consisting of a DINOv3 backbone \citep{simeoni2025dinov3} and a task-specific canopy height estimation head. The model is trained on a curated data set of paired LiDAR measurements and RGB imagery from both airborne and spaceborne platforms. For inference, RGB satellite imagery from the Maxar Vivid2 mosaic, consisting of WorldView-2, WorldView-3, and QuickBird II imagery acquired between 2017 and 2020, is used. The resulting CHMv2 product provides canopy heights with global coverage and an approximate spatial resolution of 1~m (about 1.2~m at the equator and approximately 0.73~m in Germany) \citep{brandt2026chmv2}.

From both canopy height maps, tiles intersecting the national boundaries of Germany were selected and organized using a GDAL tile index with EPSG:25832 as coordinate reference system. This ensures a unified configuration consistent with the data sets provided by the BKG.

\subsection{Reference data for linear woody feature validation}\label{sec:ref_data}

\begin{figure}[htbp]
\centering
\includegraphics[width=\textwidth]{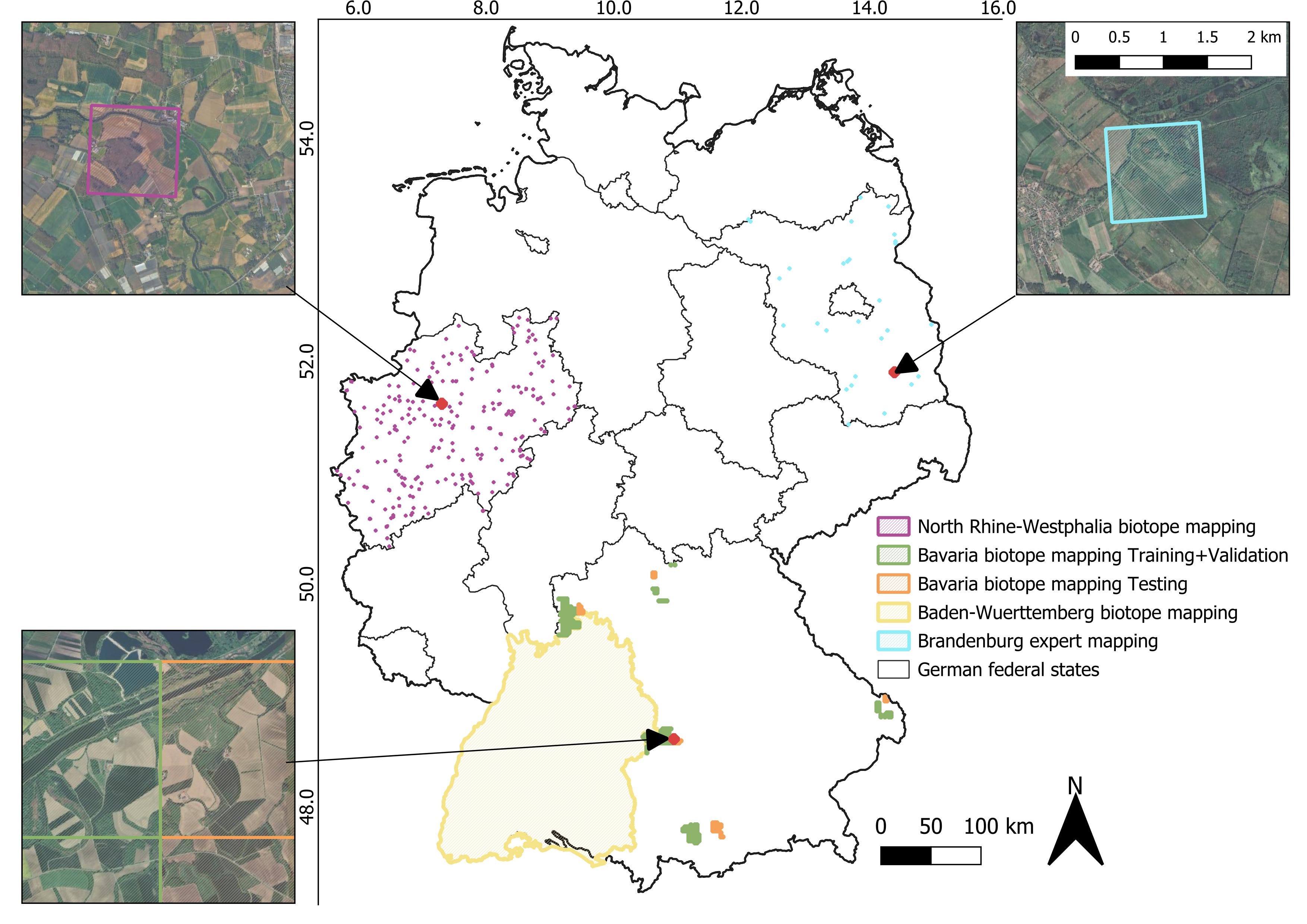}
\caption{Overview of the spatial distribution of the 5 different reference data sets. The reference data of North Rhine-Westphalia, Baden-Wuerttemberg and Brandenburg are based on federal state biotope mapping campaigns. The reference data of Bavaria has been refined by \citep{HUBERGARCIA2025101451} based on federal state data, and is presented as two separated train-val and test splits, as they have been used in the study of \citet{HUBERGARCIA2025101451}.}
\label{fig:ref_data}
\end{figure}

The reference data set was compiled with the aim of building a diverse data set in terms of spatial distribution over Germany, and spatial coverage versus label quality to perform a comprehensive evaluation of the derived linear woody vegetation features, see Figure \ref{fig:ref_data}. To this end, five independent data sources from different federal states in Germany were used.

First, data from the biodiversity monitoring programme (ökologische Flächenstichprobe) in North Rhine-Westphalia were used. This data set consists of randomly distributed 1 $\times$ 1~km grid tiles in which all linear woody biotopes are comprehensively delineated, providing a high-quality and systematically sampled reference \citep{komanns2025oefs}.

Second, biotope mapping data from the federal state of Brandenburg were included \citep{LUGV2013}. As parts of this data set are up to 15 years old and linear woody landscape elements are classified into heterogeneous categories, additional processing was required. Therefore, 30 randomly selected 1~$\times$~1~km tiles were manually revised, with all relevant polygons reviewed and corrected to only contain linear woody features based on aerial imagery.

Third, biotope mapping data from Bavaria, provided by the Bavarian Environment Agency (LfU) were used. This data set contains 340 2 $\times$ 2~km tiles from five districts surveyed in 2018 to 2020. As the Bavarian mapping focuses primarily on ecologically valuable hedgerow biotopes rather than capturing all linear woody vegetation, the data set was manually complemented by \citet{HUBERGARCIA2025101451} through detailed aerial image interpretation to ensure completeness and comparability with the other reference sites. 

Finally, freely available reference data from the federal state of Baden-Wuerttemberg were incorporated to further extend the geographic coverage and increase the variability of landscape conditions \citep{ GeoportalBW}. Biotopes were mapped into two main categories: open land habitat mapping and forest biotope mapping. For open land habitat mapping, we included biotopes of the following types: field hedgerows and groves, as well as near-natural marshland forests. For forest biotope mapping, we included the biotopes field hedgerows and groves, and dwarf, stunted woody vegetation (“Krummholz scrub”). In contrast to the other reference data sets, which are characterized by high-quality labels but sparse spatial coverage, the Baden-Wuerttemberg data set was selected to provide extensive spatial coverage, while accepting a certain degree of ambiguity in the reference data where biotope mapping data might be outdated or a slight mismatch can be observed comparing reference data and aerial imagery.

\section{Method
}
\begin{figure}[htbp]
\centering
\includegraphics[width=\textwidth]{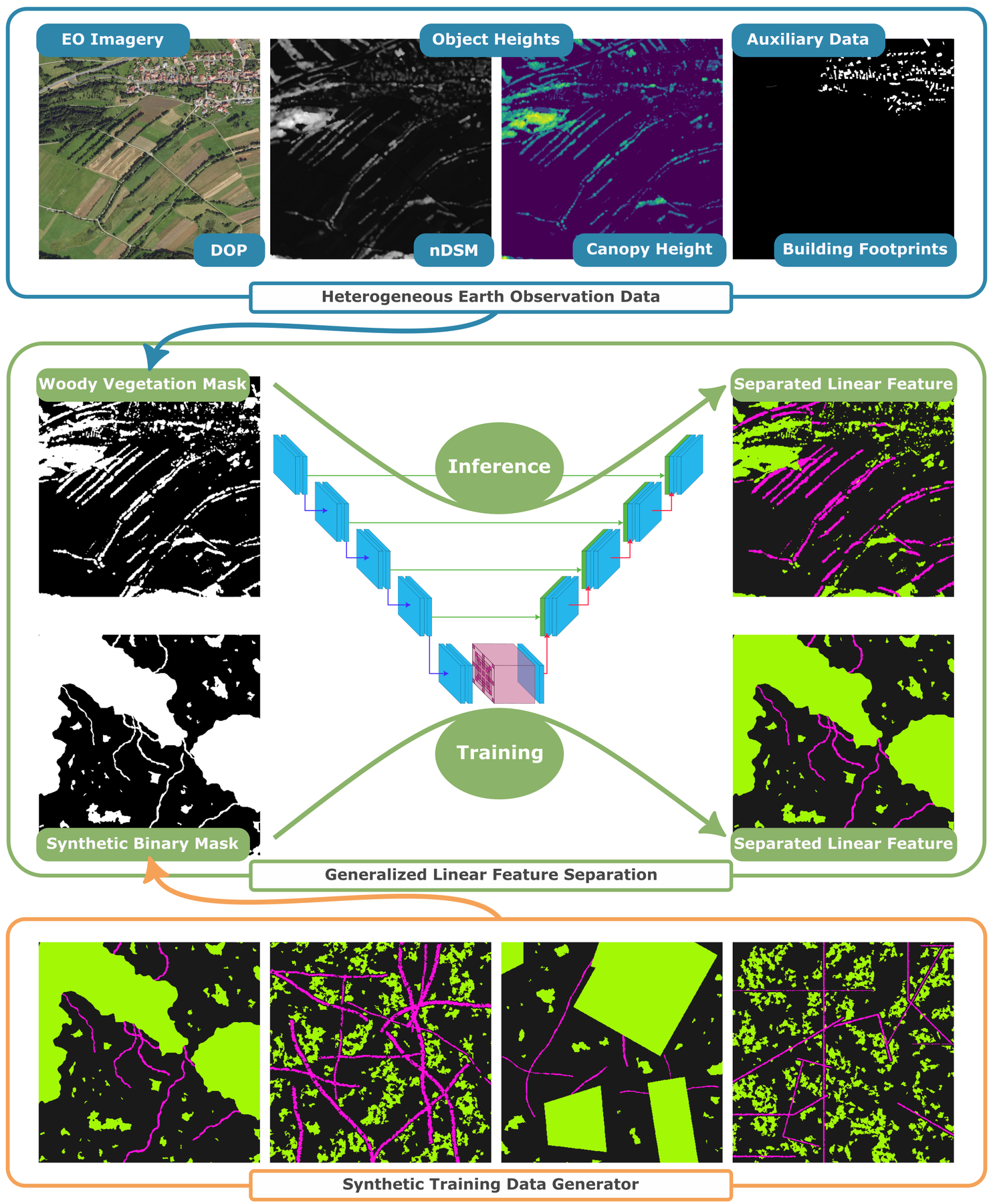}

\caption{Conceptual overview of the proposed workflow for linear woody feature mapping. The workflow consists of two major modules, a heterogeneous input data interface that consolidates Earth observation and auxiliary spatial data into a common intermediate binary woody vegetation mask, and a deep neural network trained exclusively on synthetically generated examples to separate linear from non-linear, patchy elements of this mask. By decoupling vegetation mask creation from linear feature separation, the workflow is designed to be modular and independently optimizable which is input data-independant, reusable, and scalable across different spatial contexts. DOP: Geobasisdaten: ©~GeoBasis-DE~/~BKG (2023)}
\label{fig:workflow}
\end{figure}

A central contribution and motivation of this study is to introduce a workflow, see Figure \ref{fig:workflow} and Figure \ref{fig:diff_eo_data}, which is designed to use heterogeneous input EO and other auxiliary spatial data to consolidate them individually to a common intermediate data product, a binary woody vegetation mask. These data products are then used as input into a deep learning model which is trained to separate linear features from non-linear or \textit{patchy} features, focusing on solving a generalised morphological segmentation problem rather than a data-specific appearance-based problem.

This workflow combines a flexible, customizable input data interface, until the binary woody vegetation mask, with a deep learning model that focuses on the feature's morphology and spatial context. Together, these components allow users to optimize woody vegetation mask generation and produce high-resolution, high-quality masks from the data they have available. At the same time, the deep learning model keeps its focus on separating linear woody features from complex binary masks.

The motivation for this design is rooted in building a scalable and reusable workflow. It revisits research that demonstrated how heterogeneous data has been used to derive binary woody vegetation masks for woody feature and linear woody feature extraction \citet{MAACK2017118, Meneguzzo2013, pujar2014}, and extends these workflows by providing an answer to the question of how the morphological disentanglement of complex woody vegetation masks can be addressed through a deep learning approach.

Recent studies have applied deep learning models to directly segment woody features and linear woody features in EO data \citep{HUBERGARCIA2025101451, lucas2025mapping, MURO2025114870}. While these approaches achieve strong results, they remain limited to specific spatial scales and test sites, or face challenges in scaling to larger regions due to training data availability. We argue that a data-independent and globally scalable deep learning approach would benefit when approaching it from a different perspective. Rather than matching sensor-specific appearances based on textural and radiometric details, which become increasingly difficult to obtain consistently when scaling across space and time, the model should be optimized for a more generalized task. We formulate this task to be morphological disentanglement of complex vegetation masks.

\subsection{Woody vegetation mask processing from BKG data}

The process for deriving the intermediate binary woody vegetation mask depends on the characteristics of the available input data. The three input data sets introduced in this study for deriving linear woody features can be grouped into two categories: The two canopy height map products from \citet{brandt2026chmv2} and \citet{doi:10.1126/sciadv.adh4097} require only lightweight thresholding to generate the intermediate woody vegetation mask. In contrast, the BKG data-based woody vegetation mask requires a more complex processing workflow, that explicitly accounts for factors such as acquisition date and tile-specific contrast handling, as introduced in Section \ref{sec:input_data}. The more complex processing of the BKG data demonstrates how heterogeneous input sources enable nationwide linear woody feature extraction. Using only the DOP20 data, particularly due to the large number of leaf-off acquisitions, would introduce significant challenges like multiple trained models and model retraining, or would lead to the exclusion of tiles not suitable due to their acquisition date.

\begin{figure}[htbp]
\centering
\includegraphics[width=\textwidth]{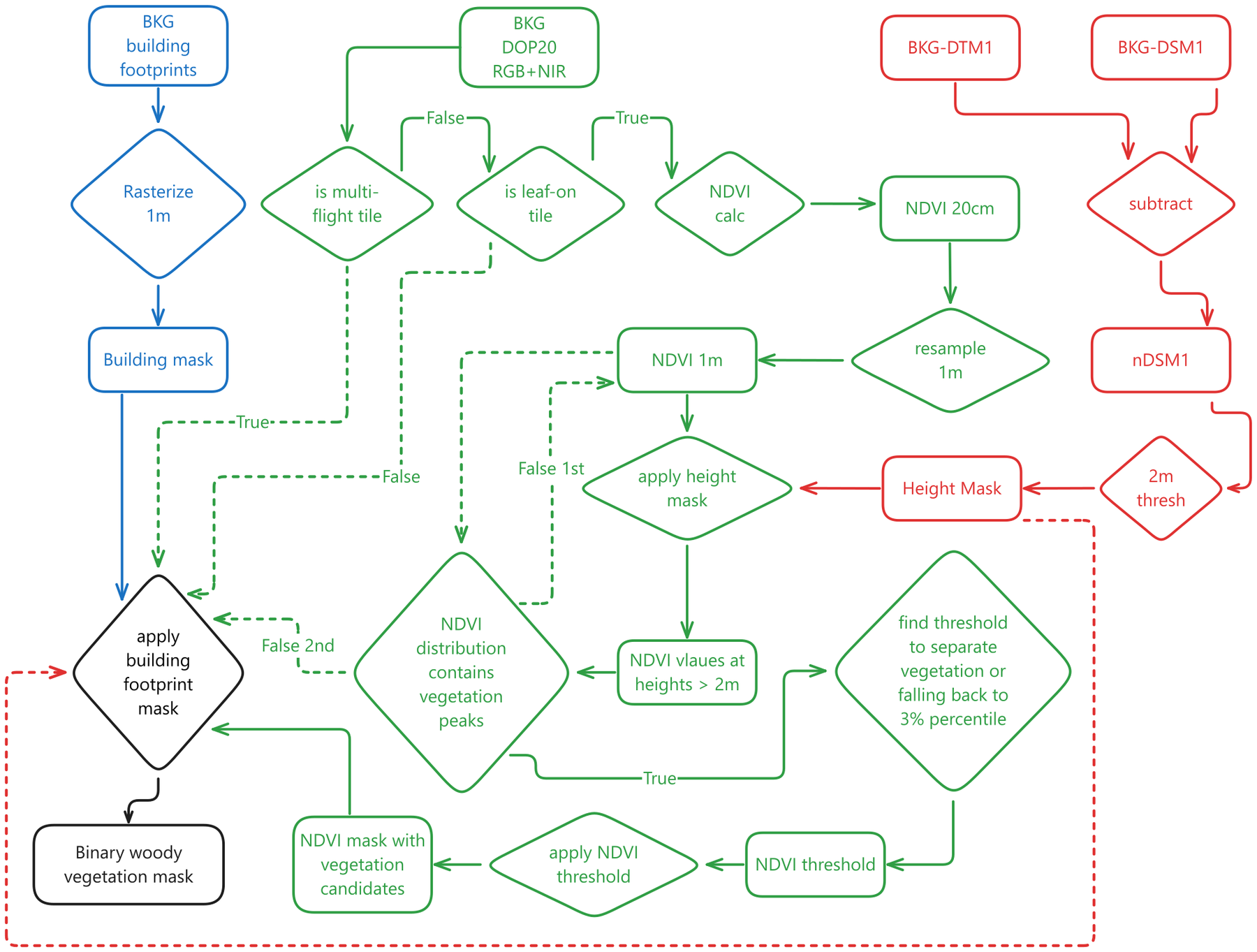}
\caption{Conditional workflow used to create binary woody vegetation masks from heterogeneous input data of the BKG data corpus, containing RGB+NIR digital orthophotos at 20~cm spatial resolution (DOP20), surface (DSM) and terrain (DTM) height models at 1~m spatial resolution, and spatial vector data of building footprints, all at a national scale for Germany.}
\label{fig:bkg_workflow}
\end{figure}

The proposed workflow for deriving the woody vegetation mask from the BKG data corpus is shown in Figure \ref{fig:bkg_workflow}. It begins with the generation of two intermediate layers, a height mask (red tiles) and a building mask (blue tiles). The DTM1 is subtracted from the DSM1 to produce a nDSM1 representing object heights independent of terrain elevation at a spatial resolution of 1~m. By applying a threshold of 2~m to this layer, a binary height mask is generated for further processing. The BKG building footprint vector data is rasterized using the nDSM1’s grid layout to produce a corresponding binary building mask.

Subsequently, the DOP20 data set is used to generate vegetation masks on a tile-by-tile basis. For each DOP20 tile, metadata are evaluated to determine whether the tile contains multiple acquisition dates or falls outside the defined leaf-on period (May 15 until September 15). In such cases, no DOP-based NDVI analysis can be conducted and the inverted building mask is combined with the height mask to derive the woody vegetation mask (red dashed arrow). If a tile contains data from a single acquisition within the leaf-on period, an additional NDVI-based masking step is applied to refine the woody vegetation mask by removing elevated non-vegetated objects not captured in the building footprint layer.

\begin{figure}[htbp]
\centering
\includegraphics[scale=0.85]{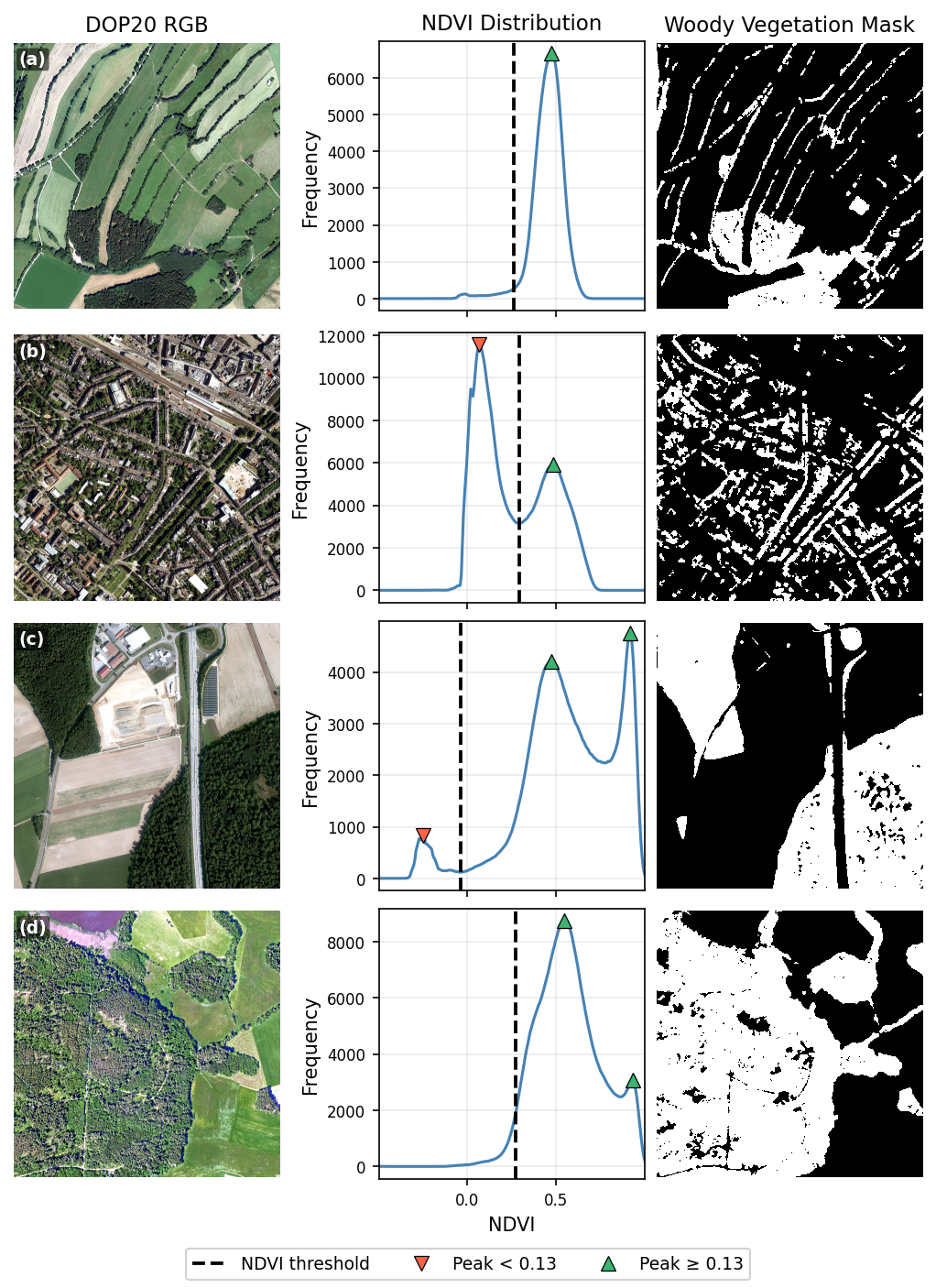}
\caption{Example DOP20 tiles under leaf-on conditions, the respective NDVI distributions with detected peaks and derived thresholds, and the resulting masks. (a) and (d) applied the 3rd percentile threshold where no peaks were detectable below an NDVI of 0.13, while (b) and (c) used the minimum between the vegetation and non-vegetation peaks to derive the threshold. DOP20: Geobasisdaten: ©~GeoBasis-DE~/~BKG (2023)}
\label{fig:ndvi_sep}
\end{figure}

Due to tile-specific contrast enhancement in the DOP20 imagery, vegetation masking must be performed individually for each tile using NDVI-based threshold detection, rather than applying a global threshold. This tile-based approach accounts for the characteristics of the DOP20 mosaic described in \ref{sec:input_data}. The NDVI is computed from the red and near-infrared bands of the DOP20 data set, and the resulting raster is resampled from 20~cm to 1~m spatial resolution to align with the height mask. To automatically determine an NDVI threshold that separates vegetation from non-vegetation for each tile individually, a distribution analysis of NDVI values is performed, see Figure \ref{fig:ndvi_sep}. Initially, the NDVI distribution is sampled by applying the height mask, thereby focusing the NDVI distribution analysis to objects exceeding 2~m in height, following \citet{MAACK2017118}. A peak detection algorithm \citet{2020SciPyNMeth} is then applied, using a prominence of 20\,\% of the histogram range (max count – min count), and a minimum peak distance of 10 bins where the underlying NDVI data is clipped to -0.5 and 1, and binned within 255 bins. Furthermore, we used an NDVI value of 0.13, still sensitive to grass land as reported by \citet{MARTINEZ2023115155}, to perform a first separation between detected peaks into vegetation and non-vegetation peaks.

For distributions with multiple peaks across both categories, the NDVI threshold is defined as the minimum value between the highest non-vegetation peak and the lowest vegetation peak. If only vegetation peaks are detected, the threshold is set to the 3rd percentile of the NDVI distribution. If no peaks or only non-vegetation peaks are identified, the NDVI distribution is sampled from the full NDVI raster. The peak detection and thresholding procedure is then repeated. If no valid threshold can still be detected after this second iteration, the final woody vegetation mask is derived using only the height mask combined with the inverted building mask, analogous to tiles acquired under leaf-off conditions or with multiple acquisition dates. This approach ensures robust handling of tiles with limited data as they appear in border regions with sparse valid observations, or for tiles showing mainly water.

For tiles with a successfully detected NDVI threshold, an NDVI mask is generated by applying the threshold to the resampled 1~m NDVI raster. The final woody vegetation mask is then obtained by combining the height mask, NDVI mask, and inverted building mask.

The intermediate woody vegetation masks are stored as BYTE GeoTIFF files and organized as GDAL tile index using EPSG:25832 as the CRS. This way we create the inference-ready intermediate data product which is a consolidation point of the overall workflow before inference is performed with the deep learning model for linear woody feature separation.

For the two canopy height map products we applied height thresholds to generate the binary woody vegetation masks. We use a threshold value of 1~m for CHMv2 \citep{brandt2026chmv2} and 2~m for \citet{doi:10.1126/sciadv.adh4097} choosing values at the lower limits of the respective canopy height map confidence levels, to maximize recall for lower vegetation which otherwise would not be included. Thresholding is performed during inference on the fly when the data is loaded from the respective GDAL tile index, see \ref{sec:inference}.

\subsection{Synthetic training data generation}
Central to the linear woody feature extraction workflow introduced in this study is a deep neural network that performs an image segmentation task. It generates a segmentation map with three classes (background, non-linear, and linear) from the input binary woody vegetation mask. The model is trained entirely on procedurally generated synthetic data instead of using real world annotated data.

The use of procedurally generated image-annotation pairs in EO has been used in applications such as offshore wind farm detection \citet{HOESER2022synteo}, wind turbine detection \citet{hoeser2022deepowt}, platform detection \citet{Spanier04032026}, or the segmentation of deficient windbreaks \citet{rs14174327}. The implementation presented in this study builds on the conceptual framework of SyntEO \citet{HOESER2022synteo}, which provides a structured approach for organizing and automating synthetic scene generation.

A core concept of SyntEO is the definition of individual scene elements as abstract representations that enable controlled random sampling within a self-regularizing parameter space. By combining scene elements and defining their spatial relationships through a set of topological rules, synthetic scenes are composed. Due to the procedural nature of this approach, each pixel can be unambiguously assigned to a specific class, allowing annotations to be derived directly from the generated scenes. This enables the creation of a scalable training data set with high annotation accuracy \citep{HOESER2022synteo}.

To represent realistic scenes and capture morphological variability, synthetic training samples are composed from a set of predefined scene elements grouped into five categories: background, linear features, and large, medium, and tiny patches. Each scene is generated by placing these elements onto a canvas of size 1,024 $\times$ 1,024 pixels following a configurable layout that controls their number, spatial arrangement, and morphological characteristics.

\begin{figure}[htbp]
\centering
\includegraphics[width=\textwidth]{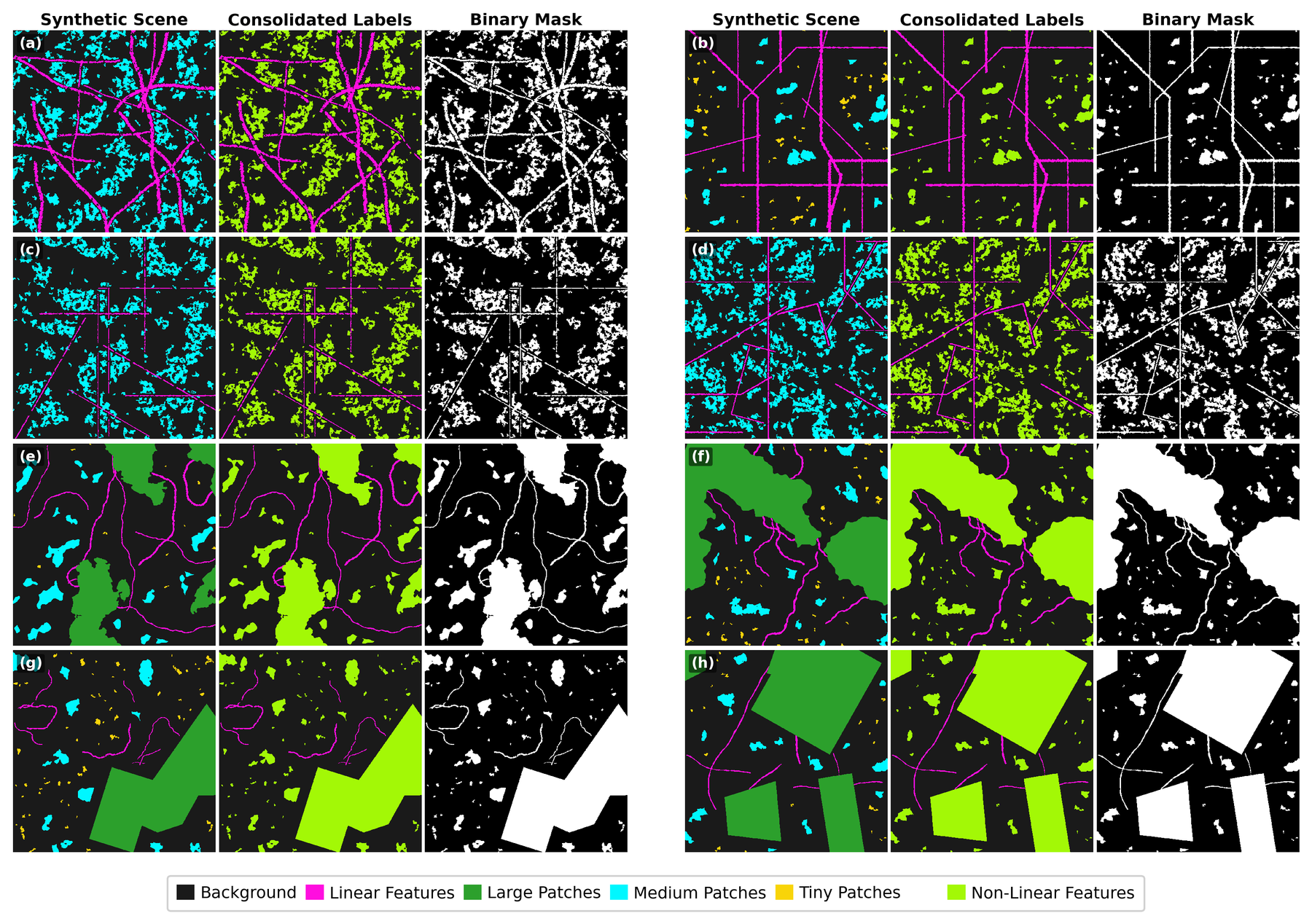}
\caption{Eight examples of synthetic scenes composed of the scene elements \textit{Linear Features, Large, Medium, and Tiny Patches} and their consolidated label masks showing \textit{Linear} and \textit{Non-Linear} features, and the fully consolidated binary masks. Different examples show different combinations of randomly sampled scene element specifications.}
\label{fig:synth}
\end{figure}

Linear features are designed to represent the two structural types \textit{angular} and \textit{organic}. Angular features consist of piecewise-straight segments (100-800 pixels) with changes in direction after each segment, based on a constrained set of angles ranging from 0 to 120 degrees (subplots b), c), and d)). In contrast, organic features follow a random-walk in which small perturbations are continuously applied to the current direction. Different curvature levels control the relationship between step size (15-20 pixels) and angular variation (10-40 degrees), resulting in shapes ranging from relatively straight (subplots a), and h)) to strongly meandering (subplots e), f), and g)), see Figure \ref{fig:synth}.

Patch elements are generated using two options. The primary option uses 2D fractional Brownian motion \citep{mandelbrot1968}, with noise and thresholding parameters controlling the size and boundary smoothness of the resulting patches. In addition, patches of size \textit{large} can be represented by randomly generated polygons (see e.g. subplots g) and h)). This allows the simulation of both irregular, natural vegetation patterns and more structured shapes, for example those typically observed within managed, agricultural landscapes with a flat terrain.

During data generation, scene configurations are sampled from predefined templates representing different landscape compositions. Within each template, parameters are randomly drawn within defined ranges while maintaining consistency between related properties (e.g., feature size and curvature). The final scene is composed by sequentially rendering background, patch, and linear elements onto a layered canvas, allowing later elements to occlude earlier ones.

Since the generated scenes contain multiple patch categories, which however are not necessary as separated patch-classes for the task at hand, they are consolidated into a single non-linear class, resulting in a final label scheme with three classes: background, linear features, and non-linear features. The corresponding input images are derived directly from the consolidated label masks by mapping foreground classes and background to binary values, to match the properties of real-world woody vegetation masks used during inference, see Figure \ref{fig:synth}. In total, the synthetic data set consists of 55,900 training examples.

\subsection{Model design and training}
\subsubsection{Neural network architecture}
\begin{figure}[htbp]
\centering
\includegraphics[width=\textwidth]{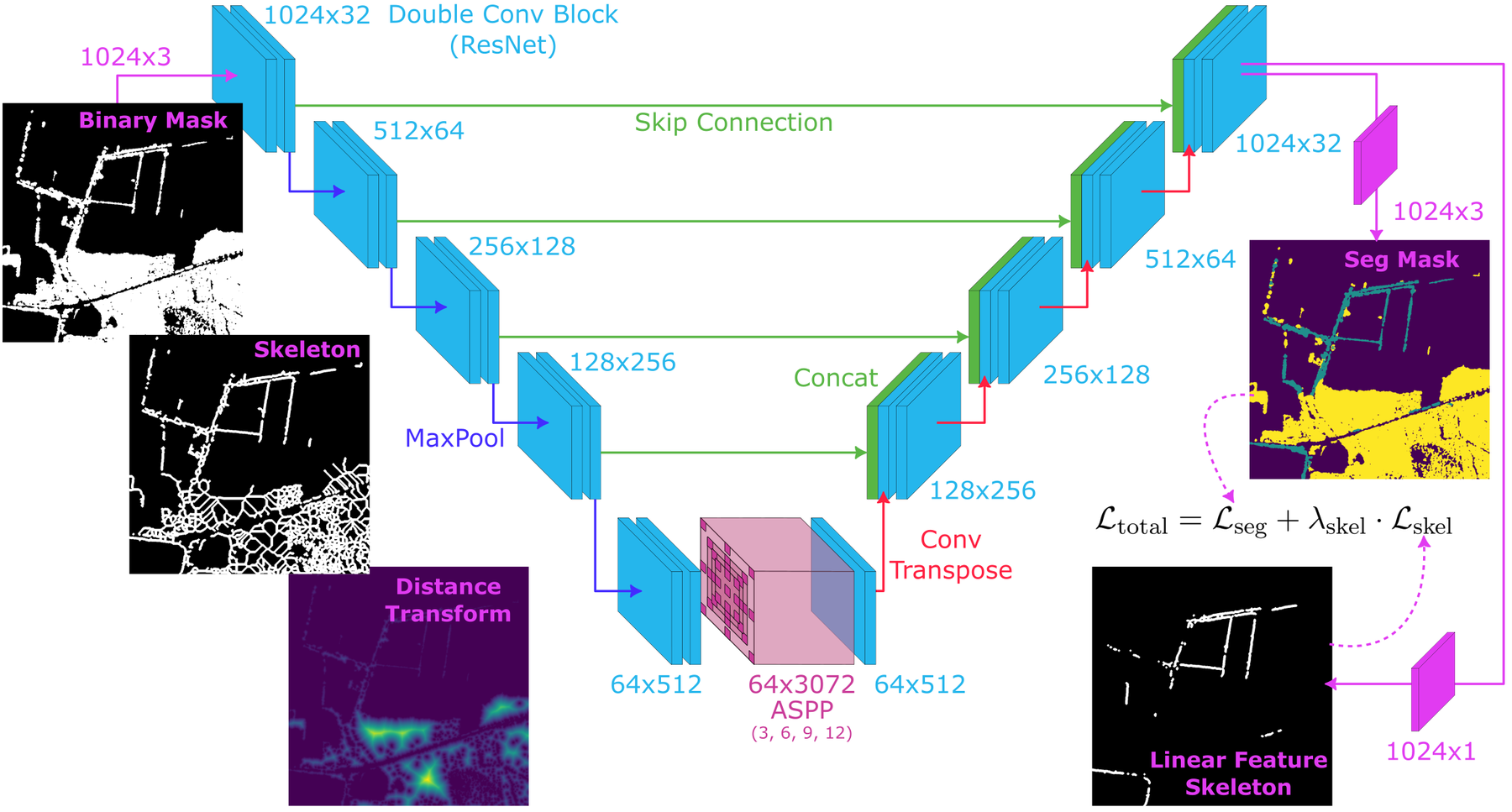}
\caption{Conceptual overview of the modified U-Net architecture with a three-channel input (binary woody vegetation mask and its corresponding skeleton and distance transform) and two output heads producing a segmentation mask and a skeleton output (the latter restricted to the linear feature class). The encoder-decoder backbone uses standard double convolutional blocks and an Atrous Spatial Pyramid Pooling (ASPP) module in the bottleneck.}
\label{fig:architecture}
\end{figure}

Linear woody feature mapping is defined as a segmentation task with three classes background, linear features, and non-linear as $\mathbf{c} = [0,1,2]$ respectively. Thereby, the target class \textit{linear features} is strongly underrepresented, as background and non-linear features naturally dominate in both real landscapes and the intermediate woody vegetation masks. Additionally, \textit{background} is already solved since the input data shows background perfectly mapped. Thus, the image segmentation task can be rendered as a separation task which aims to disentangle linear from patchy features of the positive parts of the binary input mask.
To account for the underrepresentation of the target class in the training data, focus on multi scale morphologies instead of texture, and pronounce linear signals, we customize a U-Net based architecture \citep{ronneberger2015} to shift the focus onto linear feature separation, see Figure \ref{fig:architecture}.

The base architecture is a U-Net \citep{ronneberger2015} with five double convolutional ResNet \citep{7780459} blocks in the encoder and an overall stride of 16. The bottleneck contains an Atrous Spatial Pyramid Pooling (ASPP) module \citep{7913730} with a global average pooling and 3, 6, 9 and 12 delineation rates to enhance the model’s capability to take multi scale spatial features and morphological aspects into account. Starting from the bottle neck feature depth with 512 feature maps to recover input size, transposed convolutional blocks of the decoder with skip connections to their corresponding encoder stages restore the image input size of 1,024 $\times$ 1,024 pixels. After input size is reached, two heads are configured. One to produce the three segmentation masks for background, non-linear features and linear features resulting in 1,024 $\times$ 1,024 $\times$ 3 channels and the skeleton prediction head resulting in 1,024 $\times$ 1,024 $\times$ 1 channel.

Instead of solely using the binary woody vegetation masks the input to the neural network is a three-channel input consisting of the binary woody vegetation mask and a derived skeleton \citep{vanderwalt2014scikitimage} and distance transform \citep{2020SciPyNMeth} of the entire binary mask. Especially the skeleton input layer corresponds to the skeleton prediction head whose task is to only predict these parts of the skeleton which are associated with linear features and thus forces the model to focus on these features in the shared encoder-decoder architecture. The distance transform channel introduces continuous, texture-like information to the otherwise two other, purely binary channels.

\subsubsection{Loss design}
To emphasize the linear-feature class during training, the overall loss function is defined as a weighted combination of three components:

\begin{equation}
\mathcal{L}_{\text{total}} = \mathcal{L}_{\text{wCE}} + \lambda_{\text{Dice}} \cdot \mathcal{L}_{\text{Dice}}^{(c=1)} + \lambda_{\text{skel}} \cdot \mathcal{L}_{\text{skel}},
\end{equation}

where $\mathcal{L}_{\text{wCE}}$ is a weighted cross-entropy loss over all three segmentation classes $\mathbf{c} = [0,1,2]$ (background, linear, non-linear), $\mathcal{L}_{\text{Dice}}^{(c=1)}$ is a Dice loss applied exclusively to the linear-feature class ($c=1$) with $\lambda_{\text{Dice}} = 0.3$, and $\mathcal{L}_{\text{skel}}$ is a binary cross-entropy loss applied to the output of the skeleton prediction head with $\lambda_{\text{skel}} = 0.5$.
The weighted cross-entropy loss is the main component of the objective and defined as

\begin{equation}
\mathcal{L}_{\text{wCE}} = -\frac{1}{N} \sum_{i=1}^{N} \sum_{c=0}^{2} w_c \cdot y_{i,c} \cdot \log(\hat{p}_{i,c}),
\end{equation}

with class weights $\mathbf{w} = [1.0,\ 50.0,\ 5.0]$ corresponding to the classes background, linear, and non-linear, respectively. This weighting scheme addresses the class imbalance in the training data and shifts the training focus towards the separation of linear features from complex and entangled morphologies.

To directly improve F1-score optimization for the linear-feature class, a class-specific Dice loss is added:

\begin{equation}
\mathcal{L}_{\text{Dice}}^{(c=1)} = 1 - \frac{2 \sum_{i} \hat{p}_{i,1} 
\cdot y_{i,1} + 1}{\sum_{i} \hat{p}_{i,1} + \sum_{i} y_{i,1} + 1},
\end{equation}

where the constant of 1 acts as a smoothing term for numerical stability. The final component is a skeleton-based loss $\mathcal{L}_{\text{skel}}$ that guides the model to learn the topological continuity of elongated, thin structures. It is implemented using PyTorch's \texttt{BCEWithLogitsLoss}, which combines the sigmoid activation and binary cross-entropy \citep{paszke2019pytorch}.

\subsubsection{Training setup}
The synthetic data set was split into training and validation subsets using an 80/20 split, resulting in 44,720 training samples and 11,180 validation samples. With a batch size of 32, 44,704 samples are processed per epoch. During training, metrics were tracked every 250 batches by averaging over the preceding 250 batches, alongside full validation runs.

Data augmentation included horizontal and vertical flipping, as well as random rotation with additional shift and scale variations of up to 10\,\%. Model optimization was performed using the AdamW optimizer \citep{loshchilov2019adamw}. The initial learning rate was set to 0.001 and decayed to zero using cosine annealing \citep{loshchilov2017sgdr} over 44,704 steps, corresponding to 32 epochs.

Due to the fully synthetic nature of both training and validation data, the early stopping implementation was not based on the total loss but on the validation-F1 score of the linear-feature class. This choice was motivated by observations during initial experimentation runs, where we observed that the F1 score for the target class reached a plateau at high values while the overall loss continued to decrease. This indicates that further optimization primarily improves the model fit to synthetic data without yielding meaningful performance gains for the target class.

\begin{figure}[htbp]
\centering
\includegraphics[width=\textwidth]{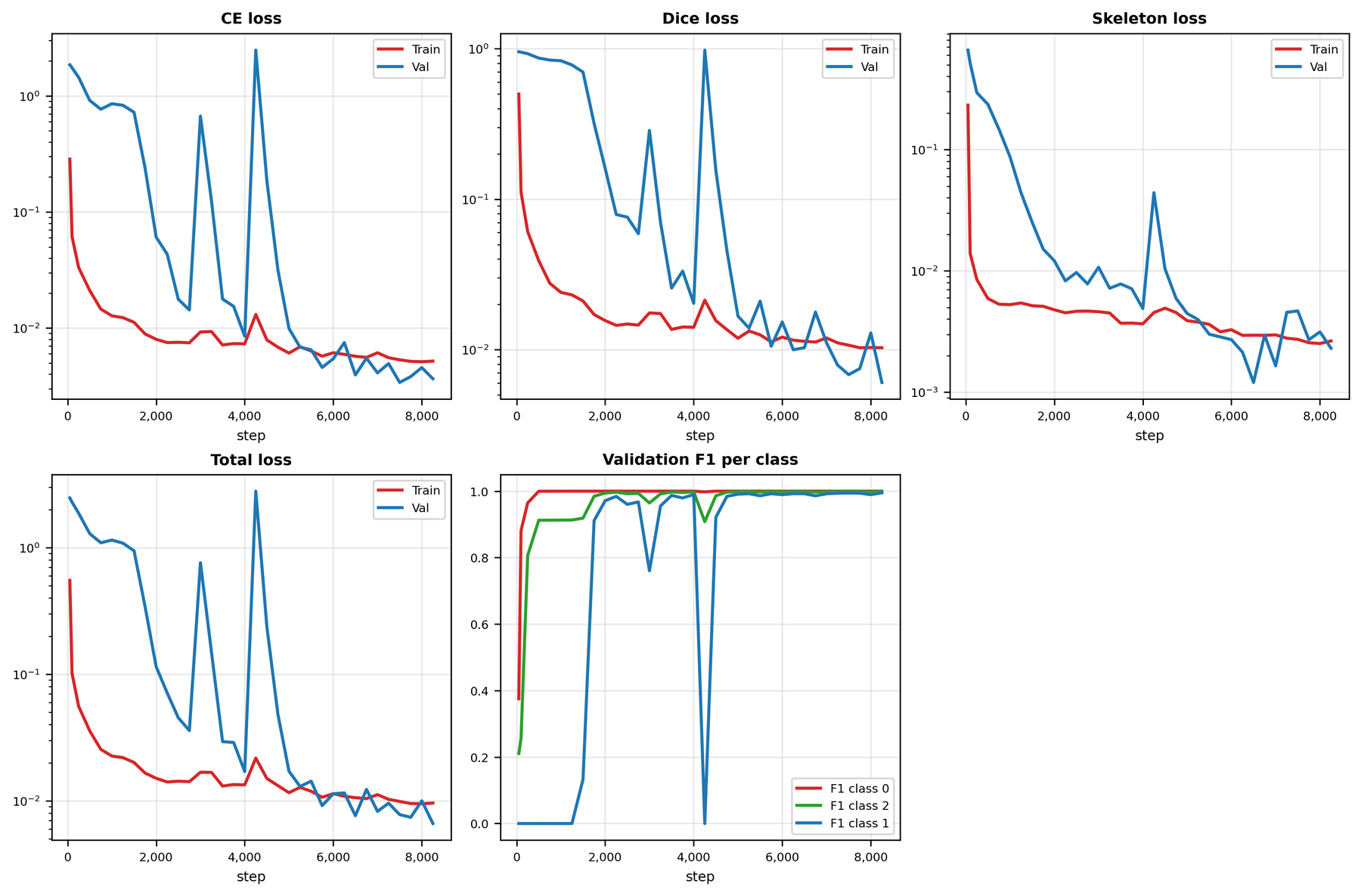}
\caption{Training loss and validation F1 scores over training steps, where each step corresponds to one processed batch. Metrics are logged every 250 steps.}
\label{fig:training}
\end{figure}

Early stopping was therefore triggered when the validation-F1 score for the linear-feature class did not improve by at least 0.01 over 2,794 training steps (equivalent to two epochs). Training stopped at step 8,250 (approximately 5.9 epochs), achieving a validation F1 score of 0.995 for the linear-feature class of the synthetic data validation split, see Figure \ref{fig:training}.

\subsection{Large-scale inference and post processing}\label{sec:inference}

For efficient and seamless inference over the nationwide binary woody vegetation masks organized as GTI, the GTI’s raster extent is used to virtually divide the mosaic into chips of 1,024 $\times$ 1,024 pixels with a horizontal and vertical overlap of 50\,\%. For each virtual chip bounds, raster values are accessed directly via the GTI and passed to the model for inference. On predictions derived from the BKG data, an additional skeleton-based refinement is applied. Those pixels that are not within a distance of 25 pixels from the predicted skeleton are removed.

To ensure seamless results, from the 50\,\% overlapping tiles, only the central 512 $\times$ 512 pixels of each prediction are used. Thereby we take edge effects into account, which are typical for convolutional neural networks and their tendency to predict with a reduced quality near tile boundaries. Furthermore, while overlapping tiles improve spatial context during inference, excluding edge regions from the final output prevents misclassification artifacts. For example, large woody patches that are only partially visible at the edge of a tile may otherwise be incorrectly classified as linear features, whereas overlapping tiles provide the correct contextual representation. Finally, these central regions are vectorized only keeping the shapes for the classes linear and non-linear features and saved to file.

This inference strategy scales effectively to EO data sets, where predictions on mosaics exceeding the input size of the neural network is a common issue. The resulting vector outputs are finally merged into a single geoparquet file per input data set to perform visualization, validation, and downstream analysis. This way we obtain three German wide results which we call ours@BKG-data, ours@Liu et al. 2023 and ours@CHMv2.

All three predictions are furthermore refined by selecting only those derived features which are inside the German national border. This is necessary for the products derived from the CHMv2 and Liu et al. 2023 canopy height maps, which originate from tiles that are not aligned with national borders. From this set of features, we filter those smaller than 250 square meters across all resolutions ranging from 0.73~m to 3~m, following \citet{HUBERGARCIA2025101451} who propose a 70 square meter threshold for an underlying spatial resolution of 20~cm. Following this step, we compile polygon features of tree-covered forest land \citep{langner2022bestockte}, bridges \citep{osm2026}, solar park areas \citep{osm2026}, and ground-mounted photovoltaic systems \citep{albert_2025_15387100} into one layer and use it to erase the segmented features. The effect, reported in Appendix A, shows marginal shifts in F1 score, but overall it decreases recall and increases precision, which was the anticipated effect of this process.

\subsection{Evaluation}

For evaluation, we use the independent reference data sets introduced in \ref{sec:ref_data}. In addition to the three data sets of linear woody features derived using the trained model, we compare our results with outputs reported in \citet{HUBERGARCIA2025101451} and \citet{MURO2025114870}. \citet{HUBERGARCIA2025101451} provide hedgerow data for Bavaria derived from 20~cm orthophotos, while \citet{MURO2025114870} present a German-wide data set derived from PlanetScope imagery at 3~m spatial resolution.

For each evaluation site, the site-specific reference area is used to clip all predicted data sets. Both reference and predicted data are then rasterized onto a common 1~m grid covering Germany. Based on these aligned raster, standard pixel-wise metrics are calculated, including precision, recall, F1 score, and intersection over union (IoU).

To complement these pixel-based metrics and account for the spatial resolution differences, we add a skeleton-based evaluation. The motivation for this lies in the fact that pixel-based IoU can be overly sensitive to thin, elongated structures, where a large proportion of pixels lie at object boundaries and small spatial misalignments lead to low pixel-wise F1 or IoU scores. This effect is particularly important to account for when comparing products derived from different spatial resolutions, such as 3~m PlanetScope-based outputs \citep{MURO2025114870} and 0.2~m resolution aerial imagery products \citep{HUBERGARCIA2025101451} as we compare here. The skeleton-based evaluation instead focuses on topological agreement over exact pixel alignment \citep{8125187, kirchhoff2024skeleton}.

Starting from the rasterized reference and predicted masks, both are thinned using skeletonization \citep{vanderwalt2014scikitimage}, yielding skeleton representations $S_{\text{GT}}$ and $S_{\text{Pred}}$. Two distance transforms \citep{2020SciPyNMeth} are then computed, $D_{\text{GT}}(i,j)$, the distance from pixel $(i,j)$ to the nearest ground truth (GT) skeleton pixel, and $D_{\text{Pred}}(i,j)$, the distance to the nearest predicted skeleton pixel. These two distance maps are then used as lookup tables to evaluate the alignment between the GT and predicted skeletons given a tolerance threshold $\tau$, which defines the maximum allowed spatial deviation. A ground truth skeleton pixel is considered a true positive if a predicted skeleton pixel lies within $\tau$ pixels. Similarly, a predicted skeleton pixel is considered a true positive if a ground truth skeleton pixel lies within $\tau$ pixels. False negatives correspond to ground truth skeleton pixels without a corresponding prediction within $\tau$, and false positives correspond to predicted skeleton pixels without a corresponding ground truth pixel within $\tau$. These definitions lead to skeleton-based precision, recall, and F1 scores, defined as:

\begin{equation}
\text{Skeleton Recall}(\tau) = 
\frac{\left|\left\{p \in S_{\text{GT}} : D_{\text{Pred}}(p) \leq \tau\right\}\right|}
{\left|S_{\text{GT}}\right|}
\end{equation}

\begin{equation}
\text{Skeleton Precision}(\tau) = 
\frac{\left|\left\{p \in S_{\text{Pred}} : D_{\text{GT}}(p) \leq \tau\right\}\right|}
{\left|S_{\text{Pred}}\right|}
\end{equation}

\begin{equation}
\text{Skeleton F1}(\tau) = 
\frac{2 \cdot \text{Skeleton Precision}(\tau) \cdot \text{Skeleton Recall}(\tau)}
{\text{Skeleton Precision}(\tau) + \text{Skeleton Recall}(\tau)}
\end{equation}

To relax the strictness of the evaluation, the threshold $\tau$ is varied over a range $\tau \in [0, \tau_{\max}]$ (default: $\tau_{\max} = 12$). The resulting metric curves are integrated to obtain a summary score:

\begin{equation}
\text{AUC}_{\text{Skeleton metric}} = 
\frac{1}{\tau_{\max}} \int_0^{\tau_{\max}} \text{Skeleton metric}(\tau)\, d\tau
\end{equation}

This area-under-curve (AUC) score provides a tolerance-aware evaluation of structural agreement of the predicted and the reference data. Metrics are computed independently for each evaluation site and are not averaged across sites, due to differences in reference data quality and definitions of linear woody features, as discussed in \ref{sec:ref_data}.

\section{Results}\label{sec:results}
\subsection{Linear Woody Features Mapping}

\begin{figure}[htbp]
\centering
\includegraphics[width=\textwidth]{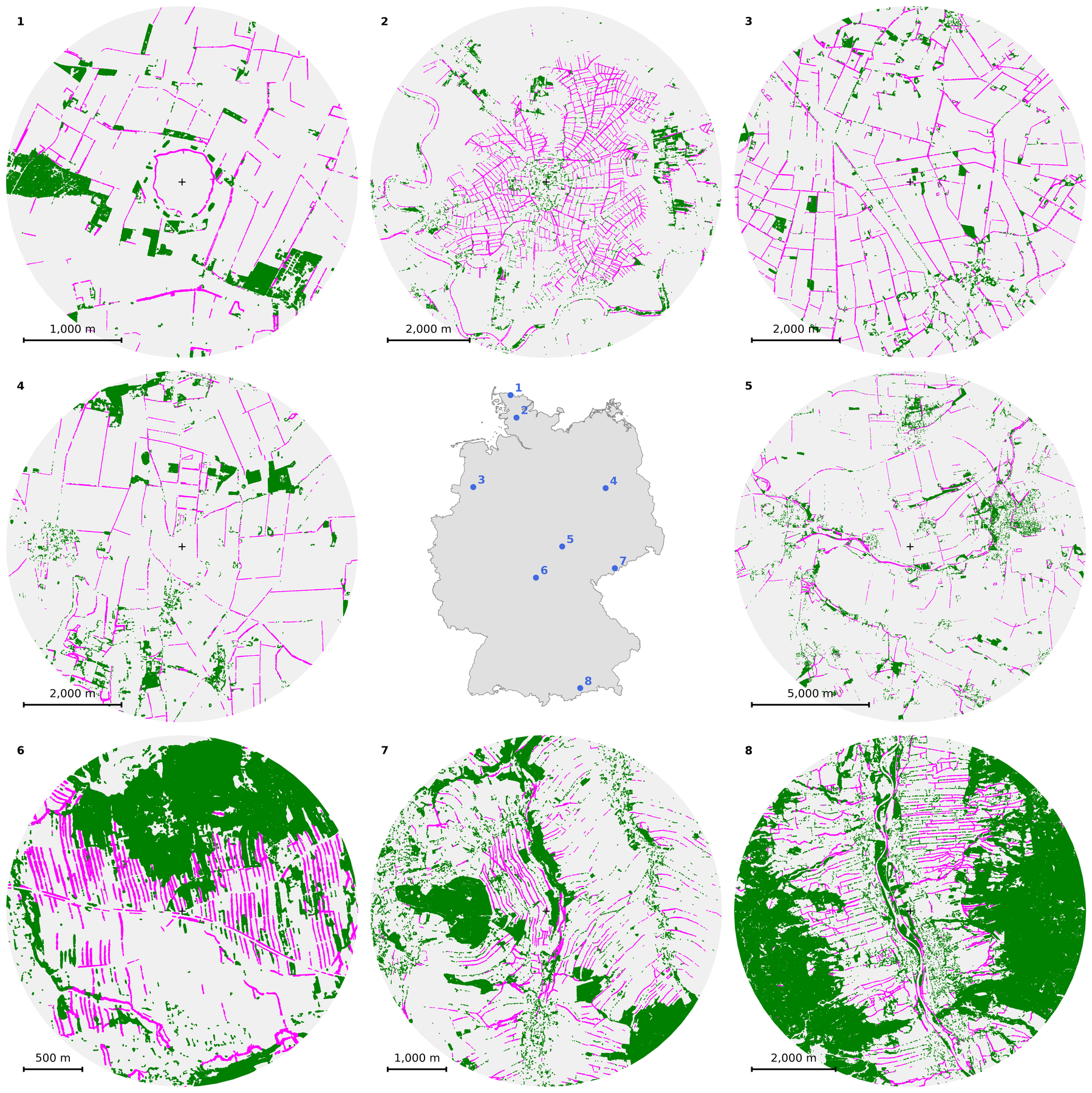}
\caption{Example prediction results derived from BKG data-based woody vegetation masks across Germany, showing a north-south gradient of diverse linear woody feature morphologies and landscapes. The gradient ranges from large-scale agricultural fields in flat terrain in the north, to high-relief terrain with greater forest cover and agroforestry ecosystems characterised by smaller field parcels in the south.}
\label{fig:result_overview}
\end{figure}

Figure \ref{fig:result_overview} provides an impression of the predictions derived from the BKG data set across Germany. These example sites follow a north-south gradient reflecting distinct landscape characteristics, where the northern sites are characterized by flat topographies with large agricultural fields, whereas sites in the central and southern parts of Germany show greater topographic relief. In these regions, agricultural land use is constrained by the terrain and forests, and linear woody features appear less frequently as long, straight elements, but rather as relatively shorter segments that conform to the underlying topography and other land use and land cover.

\begin{figure}[htbp]
\centering
\includegraphics[width=\textwidth]{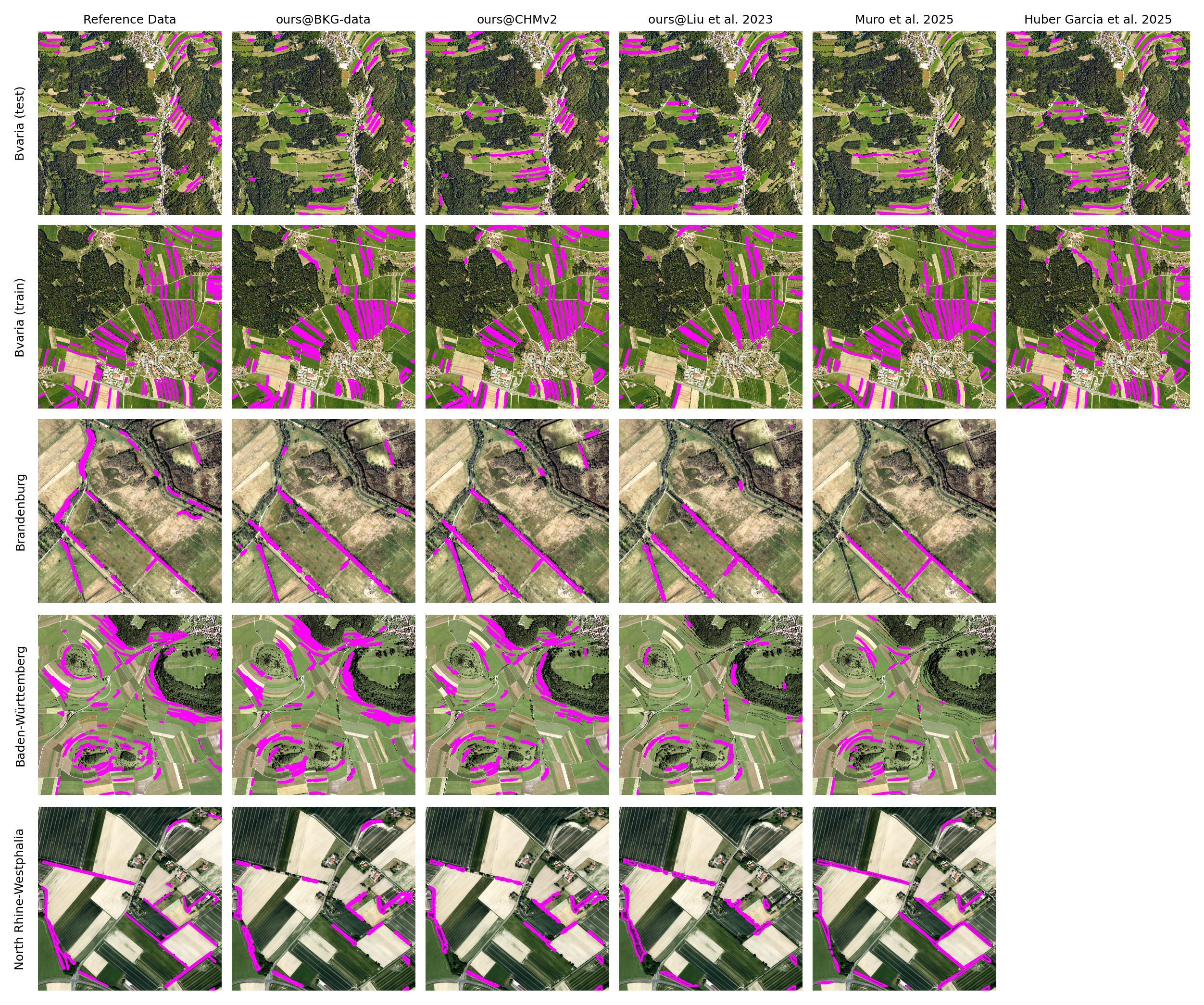}
\caption{Visual comparison of the reference data (first column) and predictions (other columns) across all evaluation sites. The predictions correspond to input data at spatial resolutions of 1~m \textit{ours@BKG-data}, 0.73~m \textit{ours@CHMv2}, 3~m for \textit{ours@}\citet{doi:10.1126/sciadv.adh4097} and \citet{MURO2025114870}, and 0.2~m for \citet{HUBERGARCIA2025101451}. Aerial Basemap: Geobasisdaten: ©~GeoBasis-DE~/~BKG (2023)}
\label{fig:eval_visual}
\end{figure}

\begin{figure}[htbp]
\centering
\includegraphics[width=\textwidth]{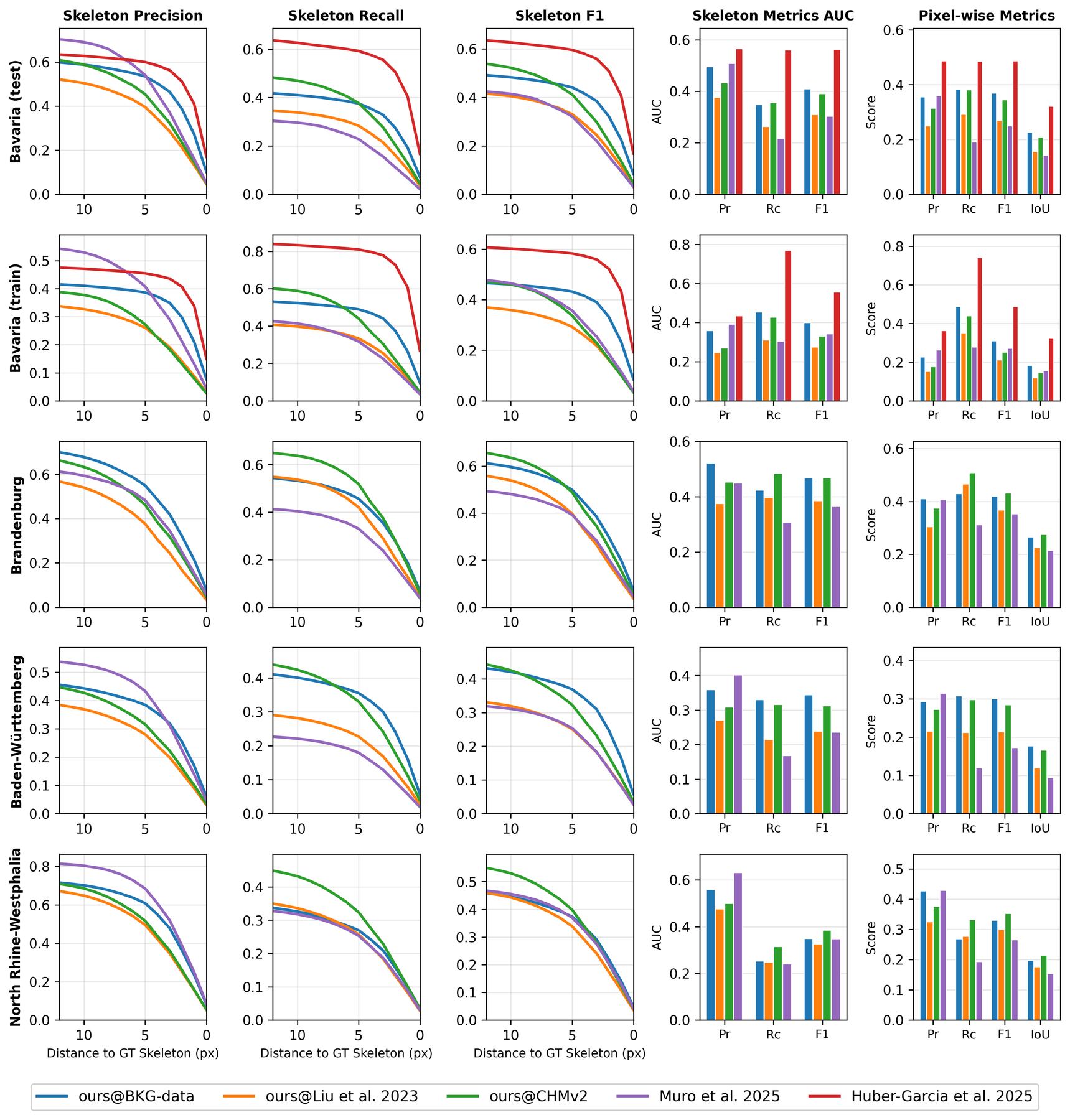}
\caption{Quantitative evaluation results across all evaluation sites and predictions. Pixel-wise metrics are reported for reference, while the evaluation is presented and discussed with a focus on skeleton-based metrics. Note that \citet{HUBERGARCIA2025101451} provide predictions only for Bavaria, and therefore no metrics are reported for other sites.}
\label{fig:eval_metric}
\end{figure}

In order to compare all three derived German-wide linear woody feature data sets and the two existing data sets with the reference data, Figure~\ref{fig:eval_visual} provides a visual impression, while Figure~\ref{fig:eval_metric} presents the quantitative comparison of all three data products and two additional existing data sets containing linear woody features in Germany.

The spatial resolution of the input data differs across results, with 1~m for \textit{ours}@BKG-data, 0.73~m for \textit{ours}@CHMv2, 3~m for \textit{ours}@\citet{doi:10.1126/sciadv.adh4097} and \citet{MURO2025114870}, and 20~cm for \citet{HUBERGARCIA2025101451}. A first visual inspection shows that results derived from input data with a spatial resolution finer than 3~m yield more detailed outputs, a finding that is supported quantitatively by the metrics presented in Figure~\ref{fig:eval_metric}.

Before presenting the quantitative results in detail, a general comparison of skeleton-based metrics with pixel-wise metrics shows that the performance advantage of results based on higher resolution input data is amplified when using pixel-wise metrics, suggesting a higher degree of alignment in the predicted masks. The effects that appear pronounced in pixel-wise F1 and IoU scores become less distinct or disappear entirely when evaluated in terms of topological correctness, as captured by the skeleton-based metrics. We therefore focus on skeleton based metrics, in order to compensate for effects coming from different resolutions as much as possible.

For the two Bavarian reference sets, representing the train and test splits used in \citet{HUBERGARCIA2025101451}, the model optimized for the underlying data as trained in \citet{HUBERGARCIA2025101451} performs best, followed by \textit{ours}@BKG-data for both the test and train splits. \textit{ours}@CHMv2 performs on par with \textit{ours}@BKG-data on the Bavarian test split, while the two 3~m-based predictions, \textit{ours}@\citet{doi:10.1126/sciadv.adh4097} and \citet{MURO2025114870}, rank lowest. For the Bavarian train split, \citet{MURO2025114870} ranks before \textit{ours}@CHMv2.

Since the study of \citet{HUBERGARCIA2025101451} focused Bavaria and data acquired at 20~cm spatial resolution during leaf-on periods, and thus linear woody features are reported only for the federal state of Bavaria, \citet{HUBERGARCIA2025101451} is not included in the following comparisons performe din other federal states. For Brandenburg, Baden-Wuerttemberg, and North Rhine-Westphalia (NRW), a consistent pattern can be observed. \textit{ours}@BKG-data and \textit{ours}@CHMv2 outperform \textit{ours}@Liu et al. 2023 and \citet{MURO2025114870} across all regions. The only exception is NRW, where \citet{MURO2025114870} reaches parity with \textit{ours}@BKG-data on the skeleton-based F1 score, and \textit{ours}@CHMv2 is the best-performing model on this specific site.

Taken together, among the models trained to provide generalised predictions at the national scale, \textit{ours}@BKG-data and \textit{ours}@CHMv2 achieve the strongest performance, with \textit{ours}@BKG-data leading overall.
 
\subsection{Overall Workflow}

The proposed workflow demonstrates advantages in reusability, optimization, and generalization. Central to the workflow is the intermediate binary woody vegetation mask, which decouples heterogeneous input data processing from the morphological separation task performed by the segmentation model. Generating three national-scale linear woody feature maps from three different input data sources using a single trained model is an important step towards a reusable mapping workflow. This makes it unnecessary to either retrain a model from scratch for each new input data configuration, or to acquire data that conforms to narrow input prerequisites of existing pretrained models, which are typically optimized for specific sensors, spatial resolutions, or seasonal acquisition conditions. Instead, a single shared model can serve a broad user community for linear woody feature mapping at national scale or for local applications similarly, if necessary, at repeated points in time.

This reusability is directly connected to the optimization potential of the workflow, which operates through two independent modules. The first module is the vegetation mask processor, which we intentionally designed as a decentralized component that enables practitioners to generate the best possible intermediate woody vegetation mask from whatever data they have available. 

Improvements to this module, whether through the addition of auxiliary data sources or refinements to the processing logic, contribute directly to the final linear woody feature product without requiring any modification to the segmentation model. This shifts parts of the optimization responsibility to domain experts in remote sensing and geospatial data processing, especially including data officers and practitioners with access to high-quality regional data sets, without requiring expertise in deep learning model training. Available research in the remote sensing domain \citep{MAACK2017118, Meneguzzo2013, pujar2014} clearly shows that proposed approaches which have worked on optimizing this module component could today be revisited and plugged into the second module in order to investigate which site-vegetation mask processor-data combination yield the best fits and might be scalable to national levels.

The second module is the centralized segmentation model itself. Improvements to this component benefit all users and applications built on the workflow. Because the model is trained only on synthetic data, its optimization is entirely decoupled from the availability of real-world annotated training data. This allows deep learning specialists to improve model performance independently, using the synthetic data generator as a scalable and adjustable optimization component, without requiring access to proprietary or regionally restricted data sets. Together, these two modules establish separated optimization pathways that distribute work to improve and advance this workflow across expert domains.

Finally, the results demonstrate that abstracting heterogeneous input data to a binary woody vegetation mask, and rendering the downstream task of linear woody feature mapping as a purely morphological separation problem solved by a model trained entirely on synthetic data, is sufficient to produce results outperforming existing national-scale linear woody feature maps.

\section{Discussion}

\subsection{Input Data Implications}

The intermediate woody vegetation mask enables flexible and heterogeneous data input, nevertheless depending on the processing logic to prepare the intermediate product, it imposes constraints to the input data.

The evaluation results clearly show that spatial resolution is majorly important for prediction quality, even when the assessment is focused on topological correctness rather than pixel-wise overlap. Although the orthophotos of the BKG data corpus have a spatial resolution of 20~cm, the intermediate woody vegetation mask is generated at 1~m, constrained by the spatial resolution of the normalized digital surface model used in the processing workflow. This points out a general property of the workflow, for which the effective resolution of the intermediate product is not necessarily the highest-resolution of the input data, but can become the resolution of the coarsest layer involved. Downstream prediction quality is therefore dependent on all inputs, and potentially most sensitive to the lowest-resolution component.

In the context of the BKG data, this trade-off is justified. Only by processing the normalized digital surface model together with the high-resolution DOP20 imagery enables a nationally scalable workflow that is still able to produce results for tiles acquired under leaf-off conditions or where acquisition dates have become ambiguous during mosaicking.

A further implication has to be pointed out, the temporal consistency across the three derived products. The CHMv2 is based on imagery acquired between 2017 and 2020, the canopy height product by \citet{doi:10.1126/sciadv.adh4097} uses PlanetScope imagery from 2019, and the BKG DOP20 mosaic spans multiple acquisition years and seasons ranging from 2016 until 2022. This temporal heterogeneity means that the three products do not represent a fully synchronous state of linear woody features across Germany. Differences in predicted feature extent may reflect temporal changes additionally to differences in input data resolution and processing. This limits the degree to which the three products can be directly compared and should be taken into account when interpreting both the visual comparison in Figure~\ref{fig:eval_visual} and the quantitative results in Figure~\ref{fig:eval_metric}.

\subsection{Evaluation}

The quality of the evaluation depends on the underlying reference data, which comes from four independent data sets that differ in spatial coverage, label recency, feature definition, and mapping methodology. Due to these differences, comparisons, especially the numeric evaluation results, have to be interpreted per-site and averaging or cross-comparison is not possible, as discussed in Section~\ref{sec:ref_data}. 

This heterogeneity in reference data reflects the current state of linear woody feature annotation in Germany, where no standardized national benchmark data set currently exists. To foster the structured optimization and comparability of generalized mapping workflows, such a benchmark data set is a necessary next step.

The skeleton-based metrics introduced in this study proved to be a valuable complement if not substitute to established pixel-wise metrics, providing a more topologically oriented and resolution-agnostic evaluation on prediction quality for thin, elongated structures. We want to highlight that the skeleton F1-curve representation provides insights which allows for more nuanced interpretation as a summary score alone. For example, at the Bavarian test site, \textit{ours}@BKG-data and \textit{ours}@CHMv2 yield similar skeleton F1 AUC values, however, their F1-curves are not perfectly aligned. The \textit{ours}@BKG-data-curve maintains a longer plateau at higher $\tau$ values and drops more steeply at lower thresholds, which shows that its predicted skeletons are geometrically closer to the reference data even where the AUC scores suggest parity with the performance of \textit{ours}@CHMv2. Such curve-level interpretations will become important as models mature and the relevant question shifts from which model achieves the higher summary score to the quality of each model.

\subsection{Known Limitations}\label{sec:limitations}

For the BKG data-based product, two main constraints influence prediction quality. First, the 1~m spatial resolution of the intermediate woody vegetation mask, combined with the 2~m minimum height threshold applied during processing, leads to the rejection of low-growing or structurally narrow linear woody features from the intermediate product. This partly explains the performance gap between the BKG data-based product and the 20~cm-based results reported by \citet{HUBERGARCIA2025101451} for Bavaria.

Another known limitation of the BKG data-based linear woody feature map is that tiles acquired under leaf-off conditions are handled by a processing branch that removes building footprints but does not apply NDVI-based vegetation masking. This way, elevated, elongated, non-vegetated structures appear in the intermediate mask and subsequently in the final prediction. Two examples are industrial-scale solar panel installations and bridge structures crossing roads or waterways. By applying the post-processing step, see Section \ref{sec:inference}, we targetedly catch these objects, motivated by the study of \citet{MAACK2017118}, who showed that removing similar classes can be beneficial. This fix could also be applied within the vegetation mask processor module before inference to clean the intermediate step. However, in this study we wanted to point out the effect of refinement during post-processing as reported in Section \ref{sec:results} and Appendix \ref{app:appA}.

\subsection{Future Perspectives}

The results and limitations discussed above point towards multiple future research directions for the proposed workflow.

For the training data set, a deeper investigation of real-world vegetation mask properties, particularly the morphological diversity of linear and non-linear woody features at different spatial scales and across different landscape types, should be at the centre of future extensions of the synthetic data generator. In order to generalize beyond Germany, data sets with global coverage such as the CHMv2 \citep{brandt2026chmv2} and the European wide canopy height map of \citet{doi:10.1126/sciadv.adh4097} provide an excellent starting point for assessing the full morphological diversity of woody features across different regions, continents and climate zones.

For quality improvement for German-wide mapping one important component is the compilation of a richer set of EO and geospatial data layers for the intermediate woody vegetation mask processor. Future developments such as the BKG digital twin initiative \citep{herbst2023digizde}, which aims to provide a nationally uniform and highly precise 3D model of Germany, have the potential to substantially improve the intermediate product quality and thus the final linear woody feature maps generated by the current workflow.

Finally, the ability to advance the field of linear woody feature mapping in a structured and comparable way depends on the availability of well-defined benchmark data sets. Currently, no such standardized data set exists for linear woody features at the national scale in Germany, or larger spatial scales like Europe or globally. Establishing a benchmark data set with clearly defined annotation protocols, representative geographic coverage, and permissive redistribution licensing would provide a common foundation and boost for future research, where currently compiling reference data sets is a considerable work package of linear woody feature studies.

Such a benchmark is furthermore a major corner stone towards a truly globally generalised model. We argue that the proposed workflow is, by its design, capable of supporting a global linear woody feature map. With high-quality canopy height products of global coverage, such as the CHMv2 \citep{brandt2026chmv2}, already integrated into this workflow, scaling to European or global linear woody feature mapping is no longer a distant vision but a concrete research objective. The major impediments we identify are a global benchmark data set, targeted synthetic data extensions, and structured experimentation to identify a globally generalised model.

\section{Conclusion}

Linear woody features such as hedgerows, tree lines and riparian woody strips provide vital ecosystem services, especially in managed agricultural landscapes, which makes their integration, preservation and restoration an important task. Automated mapping is crucial for managing these features at scale. However, at national level such as in Germany, linear woody features appear in highly variable morphological configurations. Additionally, the available very high-resolution input data are also heterogeneous in terms of acquisition dates, spatial coverages, sensor characteristics and preprocessing chains. This situation is getting even more challenging when leaving the national scale and extending these efforts to continents or even globally. Mapping linear woody features systematically under these conditions calls for a workflow which can compensate for such differences and is, as far as possible, agnostic to the specifications of the input data.

To address this, we introduce an approach that translates the task of linear woody feature mapping from EO data at national scale into a more general morphological separation problem. We rephrase it as the disentanglement of linear from non-linear features within binary woody vegetation masks. The workflow is built around two modules. The first translates heterogeneous input data into a binary woody vegetation mask, and the second is a deep neural network that performs the morphological feature separation, resulting in linear woody feature maps. To achieve complete decoupling of the two modules, the deep neural network is trained solely on synthetically generated data. This design creates two independently optimizable components. The woody vegetation mask processor offers decentralized adaptation potential, enabling a broad user community to generate high-quality masks from whatever data they have at hand, while improvements to the deep neural network and to the synthetic training data representation would strengthen a central component with the potential to improve all linear woody feature map workflows that depend on it.

The generalizability of the workflow was demonstrated by deriving three national-scale linear woody feature maps for Germany from three different input sources with a spatial resolution of 0.73~m, 1~m and 3~m. For each input, the trained model provides a national-scale linear woody feature layer within the quality limits mainly determined by the underlying spatial resolution. Evaluated against independent reference data across German evaluation sites, and compared with two existing products at national and federal state scale, the proposed workflow achieves competitive to superior results at the national level.

The proposed workflow, with its neural network trained only on synthetic data, is a promising solution to the spatial transferability limitation of established approaches. Furthermore, by demonstrating the workflow on input data from the two canopy height map subsets for Germany, the possibility opens up to now scale inference beyond the national level, as these two canopy height maps provide data up to European and even global extents. Future work should therefore focus on structured experimentation and benchmarking supported by large-scale reference data sets to enable a high quality, global scale linear woody feature layer.

\newpage

\section*{Abbreviations}

\begin{tabular*}{\textwidth}{r@{\hspace{0.8em}}l@{\extracolsep{\fill}}}
ASPP  & Atrous Spatial Pyramid Pooling \\
AUC   & Area Under the Curve \\
BKG   & Bundesamt für Kartographie und Geodäsie (Federal Agency for Cartography and Geodesy) \\
CHMv2   & Canopy Height Map version 2, \citet{brandt2026chmv2} \\
CLMS   & Copernicus Land Monitoring Service \\
CRS   & Coordinate Reference System \\
DOP   & Digital Orthophoto \\
DSM   & Digital Surface Model \\
DTM   & Digital Terrain Model \\
EO    & Earth Observation \\
FN    & False Negative \\
FP    & False Positive \\
GDAL  & Geospatial Data Abstraction Library \\
GTI   & GDAL Tile Index \\
GT    & Ground Truth\\
IoU   & Intersection over Union \\
LfU	  & Bayerisches Landesamt für Umwelt (Bavarian Environment Agency) \\
LiDAR & Light Detection and Ranging \\
LISS  & Linear Imaging and Self Scanning Sensor \\
NDVI  & Normalized Difference Vegetation Index \\
NIR   & Near-Infrared \\
nDSM  & Normalized Digital Surface Model \\
OBIA   & Object-Based Image Analysis \\
RGB   & Red, Green, Blue \\
SWF	  & Small Woody Features \\
ToF   & trees outside forests \\
TP    & True Positive \\
\end{tabular*}

\section*{Author contributions}
Conceptualization TH;
Data curation TH, SA, VHG, and UG;
Formal analysis TH;
Funding acquisition CK, and UG;
Methodology TH;
Project administration CK and UG;
Software TH;
Supervision CK;
Validation TH;
Visualization TH and SA;
Writing – original draft TH, VHG, SA;
Writing – review and editing TH, VHG, SA, UG, and CK.

\section*{Competing interests}
The contact author has declared that none of the authors has any competing interests.

\section*{Acknowledgements}
The authors gratefully acknowledge the provision of the BKG data for federal authorities and authorised users pursuant to V GeoBund Geobasisdaten. The authors gratefully acknowledge the computational resources provided through the joint high-performance data analytics (HPDA) project "terrabyte" of the German Aerospace Center (DLR) and the Leibniz Supercomputing Center (LRZ). Furthermore, we gratefully acknowledge the provision of requested data to perform inference, and evaluation. LfU provided data on Biotope mapping (© Bayerisches Landesamt für Umwelt), LANUK NRW provided data on the "Ökologische Flächenstichprobe" (© Landesamt für Natur, Umwelt und Klima Nordrhein-Westfalen), and Liu et al. 2023 provided the German wide canopy height map at a 3~m resolution used for inference. Muro et al. 2025 provided the German wide hedgerow map at a 3~m resolution used for comparison. Indicated OSM feature data copyrighted OpenStreetMap contributors and available from https://www.openstreetmap.org.

\section*{Funding}
This research did not receive any specific grant from funding agencies in the public, commercial, or not-for-profit sectors.

\section*{Declaration of generative AI and AI-assisted technologies in the manuscript preparation process}
We acknowledge the use of DeepL, which includes a generative AI for English language translation and editing. All AI-generated text suggestions have undergone rigorous revision by the authors.

\bibliographystyle{abbrvnat}
\bibliography{references}

\newpage
\appendix

\section{Appendix}

\subsection{Post-processing report}
\label{app:appA}

\begin{figure}[htbp]
\centering
\includegraphics[width=\textwidth]{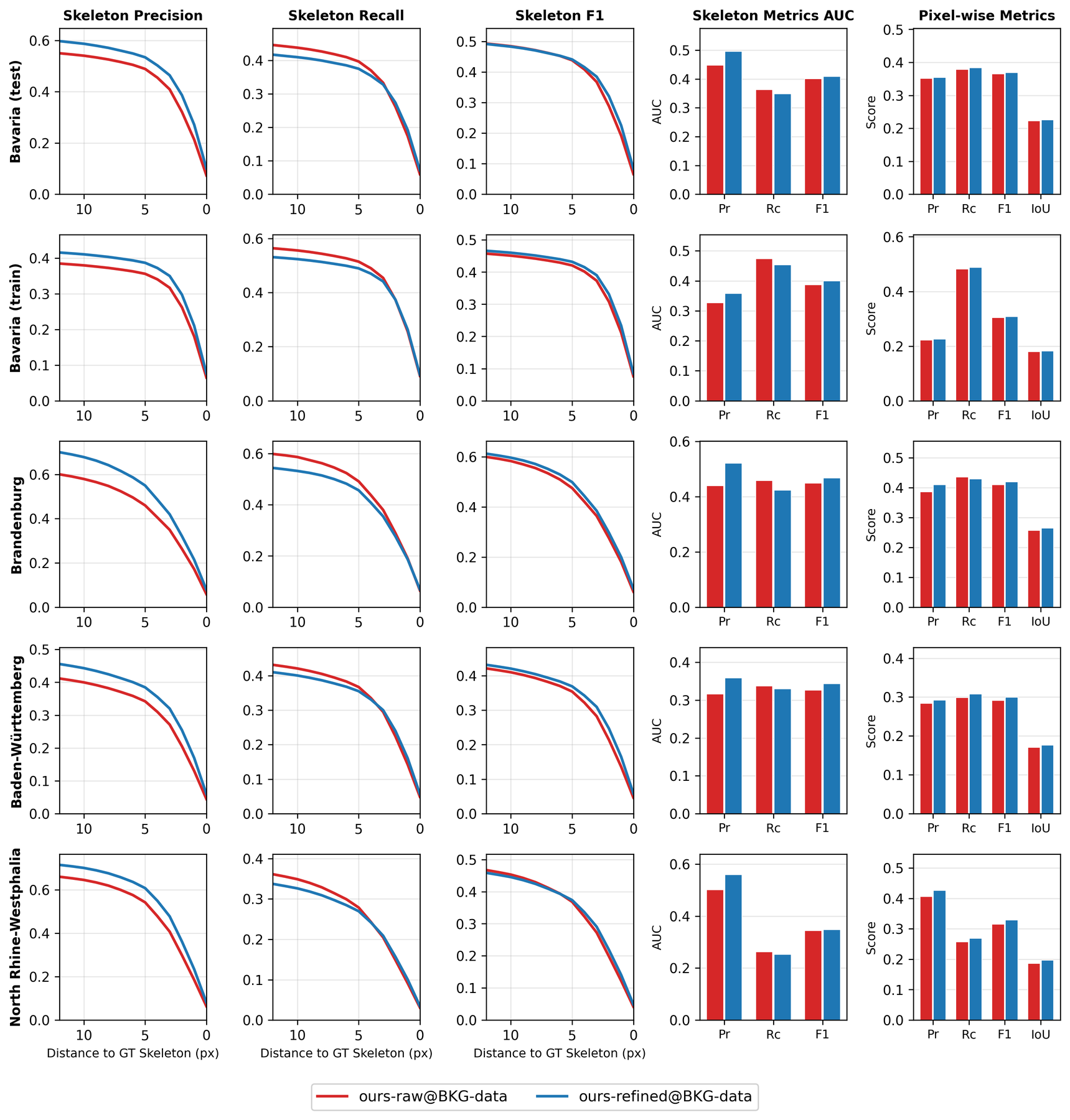}
\caption{Comparison of the evaluation metrics for the BKG data-derived linear woody feature data product before and after the refinement step described in Section \ref{sec:inference}.}
\end{figure}

\begin{figure}[htbp]
\centering
\includegraphics[width=\textwidth]{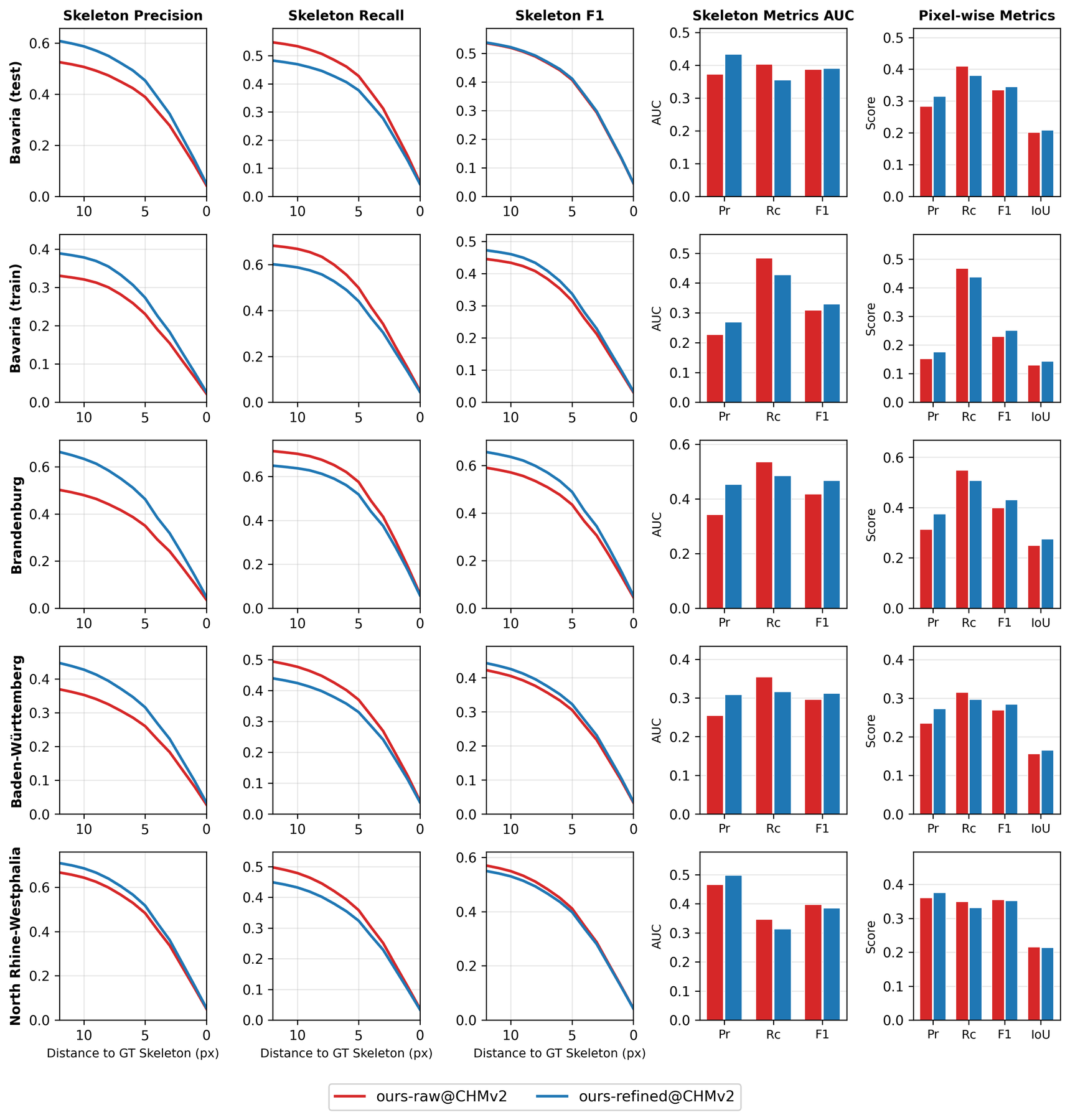}
\caption{Comparison of the evaluation metrics for the CHMv2 derived linear woody feature data product before and after the refinement step described in Section \ref{sec:inference}.}
\end{figure}

\begin{figure}[htbp]
\centering
\includegraphics[width=\textwidth]{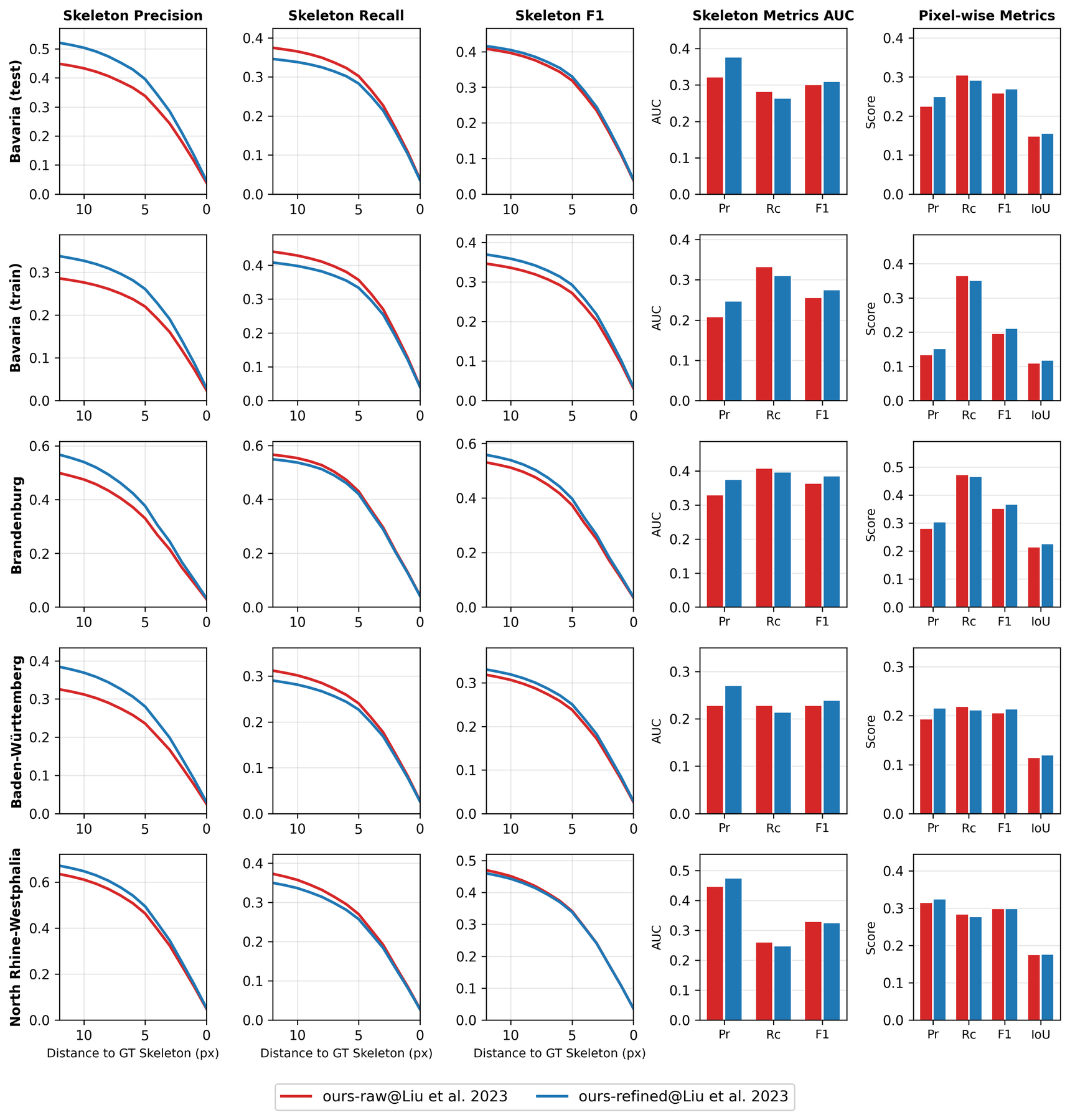}
\caption{Comparison of the evaluation metrics for the \citet{doi:10.1126/sciadv.adh4097} derived linear woody feature data product before and after the refinement step described in Section \ref{sec:inference}.}
\end{figure}

\end{document}